\documentclass{article} % For LaTeX2e
\usepackage{iclr2025_conference,times}

\usepackage{times}
\usepackage{latexsym}

\usepackage[T1]{fontenc}
\usepackage[utf8]{inputenc}

\usepackage{inconsolata}

\usepackage{graphicx}

\usepackage{microtype}
\usepackage{subfigure}
\usepackage{booktabs} % for professional tables
\usepackage{colortbl}
\usepackage{makecell}
\usepackage{array}      % For >{\bfseries}r new column type
\usepackage{svg}

\usepackage{pifont}
\usepackage[dvipsnames,table]{xcolor} % 柔和配色（含表格）
\usepackage{siunitx}
\usepackage{multirow}

\usepackage[most]{tcolorbox}
\usepackage{xcolor}
\usepackage{fontawesome5} % 如果您想加图标 (可选)

\definecolor{breadthorange}{RGB}{255, 127, 14} % 主橙色
\definecolor{breadthbg}{RGB}{255, 248, 240}    % 极淡橙背景
\definecolor{depthblue}{RGB}{31, 119, 180}     % 主蓝色
\definecolor{depthbg}{RGB}{245, 250, 255}      % 极淡蓝背景
\definecolor{codebg}{RGB}{245, 245, 245}       % 灰色背景

\newtcolorbox{promptbox}[2][]{%
    breakable,                 % 允许跨页
    enhanced,                  % 启用高级绘图
    colback=codebg,            % 背景色
    colframe=gray!60,          % 边框色
    coltitle=black,            % 标题文字色
    fonttitle=\bfseries\sffamily,
    title={#2},                % 标题参数
    arc=2pt,                   % 圆角半径
    boxrule=0.8pt,             % 边框粗细
    left=6pt, right=6pt, top=6pt, bottom=6pt, % 内部边距
    fontupper=\small\ttfamily, % 框内字体 (打字机字体)
    #1                         % 可选参数
}

\newtcolorbox{breadthprompt}[2][]{%
    breakable,
    enhanced,
    colback=breadthbg,
    colframe=breadthorange,
    coltitle=black,
    fonttitle=\bfseries\sffamily,
    title={\faFire~#2},        % 标题带图标 (需要 fontawesome5)
    attach boxed title to top left={yshift=-2mm, xshift=2mm}, % 标题悬浮效果
    boxed title style={
        colback=breadthorange, 
        sharp corners=downhill, 
        arc=2pt,
        frame hidden
    },
    separator sign=none,
    arc=3pt,
    boxrule=1pt,
    top=12pt,                  % 顶部留出空间给悬浮标题
    fontupper=\small\ttfamily,
    #1
}

\newtcolorbox{depthprompt}[2][]{%
    breakable,
    enhanced,
    colback=depthbg,
    colframe=depthblue,
    coltitle=black,
    fonttitle=\bfseries\sffamily,
    title={\faSearch~#2},      % 标题带图标
    attach boxed title to top left={yshift=-2mm, xshift=2mm},
    boxed title style={
        colback=depthblue, 
        sharp corners=downhill, 
        arc=2pt,
        frame hidden
    },
    separator sign=none,
    arc=3pt,
    boxrule=1pt,
    top=12pt,
    fontupper=\small\ttfamily,
    #1
}
\usepackage{hyperref}

\usepackage{enumitem} % 用于紧凑列表
\usepackage{CJKutf8}

\newcommand{\jap}[1]{\begin{CJK*}{UTF8}{min}#1\end{CJK*}}
\definecolor{successgreen}{RGB}{46, 139, 87}
\definecolor{failred}{RGB}{178, 34, 34}
\definecolor{breadthorange}{RGB}{255, 246, 235} % 更淡的橙色用于大面积背景
\definecolor{depthblue}{RGB}{240, 248, 255}   % 更淡的蓝色用于大面积背景
\definecolor{headergray}{RGB}{245, 245, 245}

\newcommand{\cmark}{\textcolor{successgreen}{\ding{51}}}
\newcommand{\xmark}{\textcolor{failred}{\ding{55}}}
\newcommand{\modelbadgel}[1]{\colorbox{gray!10}{\textsf{\scriptsize #1}}} 
\newcommand{\modelbadgeh}[2]{\colorbox{#1}{\textsf{\scriptsize \textbf{#2}}}}

\newcommand{\badge}[2][blue!8]{\colorbox{#1}{\textsc{#2}}}
\newcommand{\rowindent}{\hspace*{.5em}}
\newcommand{\huggingface}{\includegraphics[height=1em]{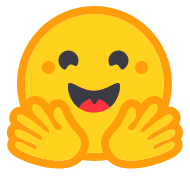}}
\newcommand{\github}{\includegraphics[height=1em]{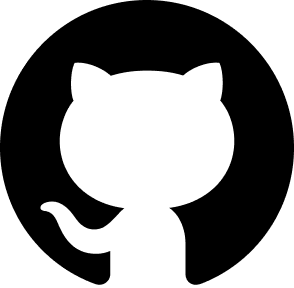}}
\usepackage{amsmath}
\usepackage{amssymb}
\usepackage{mathtools}
\usepackage{amsthm}
\usepackage{tabularx}

\usepackage{wrapfig} 

\newcommand{\ie}{\textit{i}.\textit{e}.}
\newcommand{\eg}{\textit{e}.\textit{g}.}
\newcommand{\bcot}{\textsc{B-CoT}}
\newcommand{\dcot}{\textsc{D-CoT}}

\usepackage[capitalize,noabbrev]{cleveref}

\theoremstyle{plain}

\theoremstyle{definition}

\theoremstyle{remark}

\title{Beyond Length Scaling: Synergizing Breadth and Depth for\\ Generative Reward Models}

\author{
Qiyuan Zhang$^{\spadesuit\diamondsuit*}$, Yufei Wang$^{\diamondsuit\dag}$, \\Tianhe Wu$^{\spadesuit}$, Can Xu$^{\diamondsuit\dag}$, 
Qingfeng Sun$^{\diamondsuit}$, Kai Zheng$^{\diamondsuit}$ Xue Liu$^{\heartsuit}$, Chen Ma$^{\spadesuit\dag}$\\
\vspace{2mm}
\textbf{$^\spadesuit$City University of Hong Kong} \quad 
\textbf{$^\diamondsuit$Tencent Hunyuan} \quad
\textbf{$^\heartsuit$MBZUAI}
}

\begin{document}

\maketitle

\begingroup
\renewcommand\thefootnote{*}
\footnotetext{\raggedright \mbox{Work done during the internship at Tencent Hunyuan:}~\texttt{<qzhang732-c@my.cityu.edu.hk>}}
\renewcommand\thefootnote{\dag}
\footnotetext{\raggedright \mbox{Correspondence~to:~Chen~Ma~\texttt{<chenma@cityu.edu.hk>},~Yufei~Wang~\texttt{<garyyfwang@tencent.com>},~Can~Xu}\\~\texttt{<leocaxu@tencent.com>}}
\endgroup

\begin{abstract}
Recent advancements in Generative Reward Models (GRMs) have demonstrated that scaling the length of Chain-of-Thought (CoT) reasoning considerably enhances the reliability of evaluation. However, current works predominantly rely on unstructured length scaling, ignoring the divergent efficacy of different reasoning mechanisms: Breadth-CoT (\bcot, \ie, multi-dimensional principle coverage) and Depth-CoT (\dcot, \ie, substantive judgment soundness). To address this, we introduce \textbf{Mix-GRM}, a framework that reconfigures raw rationales into structured \bcot{} and \dcot{} through a modular synthesis pipeline, subsequently employing Supervised Fine-Tuning (SFT) and Reinforcement Learning with Verifiable Rewards (RLVR) to internalize and optimize these mechanisms.
Comprehensive experiments demonstrate that Mix-GRM establishes a new state-of-the-art across five benchmarks, surpassing leading open-source RMs by an average of 8.2\%. Our results reveal a clear divergence in reasoning: \bcot{} benefits subjective preference tasks, whereas \dcot{} excels in objective correctness tasks. Consequently, misaligning the reasoning mechanism with the task directly degrades performance. Furthermore, we demonstrate that RLVR acts as a switching amplifier, inducing an emergent polarization where the model spontaneously allocates its reasoning style to match task demands. The synthesized data and models are released at \huggingface\href{https://huggingface.co/collections/DonJoey/mix-grm}{Hugging Face}, and the code is released at \github\href{https://github.com/Don-Joey/Mix-GRM}{Github}.
\end{abstract}

\section{Introduction}
\label{sec:intro}

Reinforcement learning (RL) has proven to be the critical post-training mechanism for eliciting capabilities in Large Language Models (LLMs)~\citep{ouyang2022training,deepseekai2025deepseekr1,kimiteam2025kimik15scaling}. 
However, as the ambition of RL expands from single-domain optimization (\eg, math)~\citep{hung2022coderl,shao2024deepseekmath,wang2025enhancingcodellmsreinforcement} to general-purpose alignment~\citep{harrison2024rlaif,shen2025exploring}, the Reward Model (RM) faces the challenge of providing reliable feedback for increasingly complex queries from diverse, real-world scenarios~\citep{liu2025inference,li2025generalistrewardmodelsinside}.
Addressing this challenge requires a shift in RM design. Inspired by how CoT~\citep{wei2023chainofthought,yang2025demystifying} trades inference-time compute for enhanced generalization performance, the community has increasingly adopted Generative Reward Models (GRMs)~\citep{zheng2023judging,weizhe2024self,zhang2025generative}. By prompting an explicit evaluation rationale prior to conclusion, GRMs aim to transfer the robust generalization observed in CoT generation to the task of reward modeling.

Building on these successes, existing GRM methods predominantly leverage CoT by simply scaling its length~\citep{chen2025rmr1rewardmodelingreasoning,chen2025judgelrmlargereasoningmodels,zhang2025crowd}, feeding it with massive evaluation signals, such as fine-grained features~\citep{kim2024prometheus} or multi-perspective critiques~\citep{ankner2024critiqueoutloudrewardmodels}. However, prior CoT studies~\citep{sprague2025to,besta2025demystifying,wang2024mmlupro,kambhampati2024llmscantplanhelp} have established that longer CoTs do not universally guarantee performance gains; rather, the optimal structural bias diverges significantly across domains. Crucially, recent insights from test-time scaling~\citep{li202512surveyreasoning,zhang2025surveytesttimescalinglarge} provide a theory for this divergence, identifying \textit{parallel thinking} and \textit{sequential thinking} as two fundamental, orthogonal mechanisms for amplifying intelligence. 
Conceptually, reasoning-heavy tasks (\eg, math, code) necessitate sequential verification to ensure deductive rigor~\citep{wang2024math,liu2025safe,lightman2023letsverifystepstep}, whereas semantic-heavy tasks (\eg, open-ended generation) benefit from parallel exploration to ensure comprehensive coverage of diverse possibilities~\citep{zheng2025parallelr1parallelthinkingreinforcement,pan2025learningadaptiveparallelreasoning}. 

Drawing on this distinction, we argue that advancing RM requires shifting focus from merely scaling CoT length to aligning its reasoning mechanisms with task demands. Specifically, this necessitates a transition from static, one-size-fits-all CoT templates toward a \textit{mix reasoning mechanism}.
Thus, we propose \textbf{Mix-GRM}, which implements a dynamic mix reasoning mechanism within a unified reward modeling framework. Specifically, we introduce a synthesis framework that reconfigures raw, unstructured rationales into two distinct long CoTs: Breadth-CoT (\bcot) and Depth-CoT (\dcot).
To achieve this, we first decouple unstructured rationales into atomic ``Principle–Judgment–Verdict'' units. This modularity allows us to reassemble the units into syntactically unified but structurally diverse paths. To illustrate, a \bcot{ }is synthesized by the parallel aggregation of units across diverse principles (\eg, combining an `Accuracy' unit with a `Clarity' unit) to ensure coverage. Conversely, \dcot{ }extends the CoT by first performing a direct reasoning pass to solve the instruction, thereby enabling a re-evaluated judgment grounded in the generated reasoning pass to ensure soundness.
To cultivate mechanism-adaptive alignment, we construct a synergistic mixture dataset by pairing \bcot{ }with subjective preference tasks and \dcot{ }with objective correctness tasks. We first initialize the model via SFT on this mixture and subsequently optimize it through RLVR using normal RM datasets, where only final labels are available.

Comprehensive experiments across five standard benchmarks yield three critical conclusions:
(1) \textbf{Universal SOTA Performance and Downstream Utility}: \textit{Mix-GRM} establishes a new state-of-the-art, consistently surpassing strong baselines like \textit{Skywork-Reward} and \textit{FARE-8B} on general reward benchmarks. Crucially, this superiority extends to practical downstream tasks: \textit{Mix-GRM} demonstrates best-in-class utility in both Offline RL (DPO) and Test-time Scaling (Best-of-N).
(2) \textbf{Divergent Roles of Reasoning Mechanisms}:
Our analysis reveals that \textsc{B-CoT} predominantly benefits subjective preference but degrades objective correctness, while \textsc{D-CoT} excels in correctness at the cost of preference. This confirms that the efficacy of a reasoning mechanism is task-dependent.
(3) \textbf{RLVR as a Switching Amplifier}: Mixed mechanisms provide a superior base for RL. RLVR boosts \textit{Mix-GRM} by a larger margin than the \textit{Base-GRM}. Our analysis demonstrates that RLVR automatically sharpens the mechanism allocation—spontaneously converging on \textsc{B-CoT} for preference and \textsc{D-CoT} for correctness. This confirms that optimizing how a model thinks is more critical for post-training efficacy than simply scaling how long it writes.

\section{Related Work}
\label{sec:related_work}
\subsection{Generative Reward Model}
Generative Reward Models represent a paradigm shift from scalar regression to explicit reasoning. Developing alongside the prompting-based ``LLM-as-a-Judge'' paradigm~\citep{zheng2023judging}, GRMs are explicitly trained to generate natural language rationales alongside preference decisions~\citep{weizhe2024self}. Driven by the transformative success of long CoT, the research trajectory in this field has pivoted toward continuously extending the length of these rationales. To achieve this, many work leverages RL to explicitly elicit and stabilize longer CoT traces~\citep{chen2025rmr1rewardmodelingreasoning,chen2025judgelrmlargereasoningmodels,whitehouse2025j1incentivizingthinkingllmasajudge}, while complementary efforts utilize detailed rubrics/checklists to synthetically expand evaluation coverage~\citep{kim2024prometheus,liu2025openrubricsscalablesyntheticrubric,gunjal2025rubricsrewardsreinforcementlearning,viswanathan2025checklistsbetterrewardmodels}. However, while these strategies successfully scale the quantity of reasoning, they typically rely on static, task-agnostic structures, overlooking the critical nuance that the optimal reasoning mechanism is intrinsically task-dependent.

\subsection{Breadth and Depth in Chain-of-Thought}
The evolution of CoT is fundamentally characterized by the continuous exploration of diverse structures~\citep{shinn2023reflexionlanguageagentsverbal,deepseekai2025deepseekr1}. Beyond simple linear chains, frameworks such as Tree of Thoughts~\citep{yao2023tree} and Graph of Thoughts~\citep{besta2025demystifying} introduce branching and recurrent topologies, framing reasoning as a structured search over partial thoughts. Complementing these complex structures, approaches like Skeleton-of-Thought~\citep{ning2024skeleton} and Self-Consistency~\citep{wang2023selfconsistencyimproveschainthought} demonstrate the efficacy of parallel exploration, leveraging lateral breadth to enhance robustness and coverage. Collectively, these studies establish that reasoning is not structure-agnostic; rather, specific topological priors—ranging from deep sequential trees to broad parallel ensembles—are required to unlock optimal performance across distinct domains~\citep{sprague2025to}, a distinction that our work formally adapts to reward modeling.

\section{Methodology}
\label{sec:method}

\begin{figure*}[ht]
\centering
\includegraphics[width=.98\linewidth]{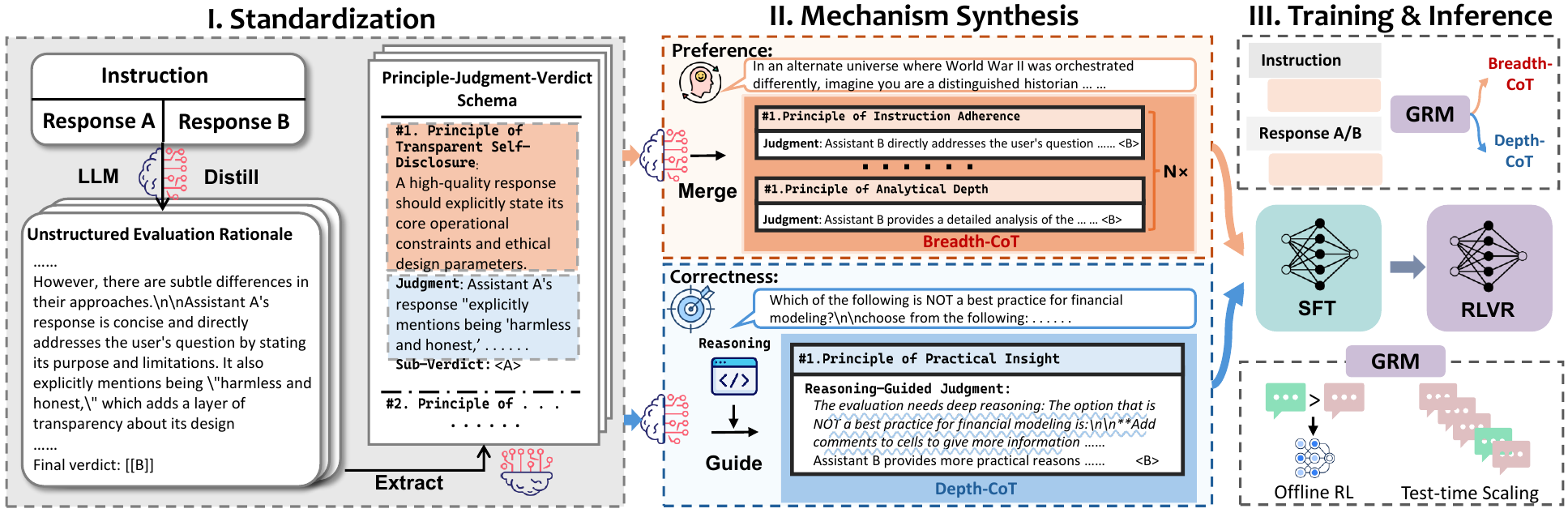}
\caption{
The pipeline of the Mix-GRM. (i) \textbf{Standardization}: We extract raw rationales into modular \textit{Principle–Judgment–Verdict} units. (II) \textbf{Mechanism Synthesis}: We reconstruct modules into \textit{\bcot} for preference or \textit{\dcot} for correctness. (III) \textbf{Training \& Inference}: Following SFT and RLVR training, the model achieves mechanism-adaptive alignment, automatically deploying the optimal mechanism for inference and providing reliable signals for downstream tasks like Offline RL and test-time scaling.
}
\label{fig:pipeline}
\end{figure*}

We propose the Mix-GRM, a framework designed to dynamically align the reasoning mechanism with intrinsic task demands. Moving beyond static, unstructured rationale sequences, our approach formalizes evaluation into two orthogonal CoTs: \textbf{\bcot}, which enforces the lateral aggregation of diverse principles, and \textbf{\dcot}, which necessitates the expansion of judgment. As illustrated in Figure \ref{fig:pipeline}, our methodology comprises three key phases: modular schema standardization (\S \ref{sec:standardization}), mechanism synthesis (\S \ref{sec:synthesis}), and mechanism-adaptive alignment (\S \ref{sec:training}).

\subsection{Problem Formulation}
\label{sec:problem_formulation}

Supposing $\{y_A, y_B\}$ denote two candidate responses generated by two assistants $A$ and $B$ for a given task instruction $x$, a normal GRM $\mathcal{M}$ produces an output sequence consisting of an explicit evaluation rationale $c$ followed by a preference verdict $v$, comparing the quality of $y_A$ and $y_B$.
\begin{equation*}
    (c, v) = \mathcal{M}(y_A, y_B \mid x). 
\end{equation*}
The objective is to ensure that the $v$ aligns with human preference. In our framework, we denote the full input triplet as $I = (x, y_A, y_B)$.

\subsection{Modular Schema Standardization}
\label{sec:standardization}

% Conventional GRMs typically produce the rationale $c$ as a free-form sequence, lacking explicit structural boundaries.
% To enable the precise manipulation of principles and judgments required for our different mechanism CoTs, we define a ``Principle-Judgment-Verdict'' Schema (Figure~\ref{fig:pipeline}, Stage I).
% We utilize a LLM to parse the raw $c$ into structured atomic units $\mathcal{S}$:
% \begin{equation*}
% \mathcal{S} = \{ (p_k, j_k, v_k) \}_{k=1}^K,
% \end{equation*}
% where $p_k$ denotes a discrete evaluation \textbf{Principle} (\eg, ``Instruction Adherence''), $j_k$ represents the specific \textbf{Judgment} (\eg, ``Response B directly addresses...'') analyzing that principle, and $v_k$ is the following \textbf{Sub-Verdict} (\eg, ``<B> is Better''). Here, a single $\mathcal{S}$ normally contains 3 to 5 units.
% Crucially, this standardization serves two strategic purposes beyond data formatting:
% (1) \textbf{\textit{Fine-grained Verifiability}}: It decouples tangled rationales into an explicit, atomic verification process. This provides the model with a much cleaner learning signal, enhancing the efficacy of SFT training.
% (2) \textbf{\textit{Syntactic Uniformity}}: It enforces a consistent format across all mechanisms, preventing the model from being confounded by stylistic variations~\citep{li2025scardataselectionstyle}. This ensures that performance gains are driven by the underlying thinking mechanisms rather than superficial surface patterns.

Conventional GRMs typically produce the rationale $c$ as an unstructured, free-form sequence.
Inspired by recent checklist-based evaluation~\citep{viswanathan2025checklistsbetterrewardmodels}, which advocates for the atomization of the complex evaluation process into checklist-driven points, we propose to reconfigure these raw rationales into a structured ``Principle-Judgment-Verdict'' Schema (Figure~\ref{fig:pipeline}, Stage I). By transforming tangled rationales into atomic units, we ensure that the RM's reasoning process is both interpretable and granularly verifiable.
Formally, we utilize a LLM to parse the raw $c$ into structured atomic units $\mathcal{S}$:
\begin{equation*}
\mathcal{S} = \{ (p_k, j_k, v_k) \}_{k=1}^K,
\end{equation*}
where $p_k$ denotes a discrete evaluation \textbf{Principle} (\eg, ``Instruction Adherence''), $j_k$ represents the specific \textbf{Judgment} (\eg, ``Response B directly addresses...'') analyzing that principle, and $v_k$ is the following \textbf{Sub-Verdict} (\eg, ``<B> is Better''). Here, $K$ typically ranges from 3 to 5.

%Compared to free-form rationales, our reconfigured CoTs provide two distinct advantages. First, the atomic decomposition of principles and judgments yields a significantly cleaner learning signal for training. Second, it enforces syntactic Uniformity across mechanisms, preventing the model from being confounded by stylistic variations~\citep{li2025scardataselectionstyle}. This isolation ensures that performance gains are driven by the underlying thinking mechanisms (\ie, Breadth vs. Depth) rather than superficial surface patterns.
This atomic decomposition yields cleaner learning signals and ensures syntactic uniformity~\citep{li2025scardataselectionstyle}, ensuring that performance gains are driven by thinking mechanisms (\ie, Breadth vs. Depth) rather than superficial stylistic patterns.

\subsection{Mechanism Synthesis}
\label{sec:synthesis}
Building on the $\mathcal{S}$, we introduce a dual-track synthesis pipeline (Figure \ref{fig:pipeline}, Stage II) to synthesize \textbf{\textsc{B-}} and \textbf{\dcot} as follows:

\paragraph{\textsc{B-CoT} Synthesis.}
We define \textsc{B-CoT} as the parallel aggregation of distinct principles, designed to overcome the narrow focus of single-pass rationale. In subjective preference tasks, where a ``good'' response is defined by the simultaneous satisfaction of multi-dimensional factors (\eg, tone, helpfulness, and creativity), single-track reasoning often fixates on dominant traits while overlooking subtle, fine-grained details.
By exploring diverse reasoning paths concurrently, parallel thinking provides a deliberative breadth that aligns with the multifaceted nature of human preference.
To simulate parallel thinking, we treat independent sampling as a stochastic exploration of the instruction's evaluative manifold. By sampling $N$ independent rationales $\{c_n\}_{n=1}^N$ from multiple cognitive trajectories, we elicit a diverse set of hidden principles that might otherwise remain dormant. These rationales are parsed into structured schemas $\{\mathcal{S}_n\}$ and subsequently unified via an LLM-based \textbf{Merge \& Deduplicate} transformation $\mathcal{T}_{\text{merge}}$:
\begin{equation*}
C_{\text{B}} = \mathcal{T}{\text{merge}} \left( \bigcup{n=1}^N { (p, j, v) \in \mathcal{S}_n } \right).
\end{equation*}
Here, we filter out lowest-frequency principles. This synthesis yields a comprehensive, non-redundant spectrum of principles, effectively expanding the model's horizontal evaluative scope.

% \paragraph{\textsc{D-CoT} Synthesis.}
% We define \dcot{ }as the expansion of judgment to ensure evaluation soundness.
% First, we prompt the LLM to generate a \textit{Reasoning Trace} $z$ derived solely from $x$, explicitly outlining logical constraints or solution paths.
% To prioritize depth under a fixed compute budget, we sample a focused subset $\mathcal{S}_{\text{sub}} \subset \mathcal{S}$ (containing at most 3 principles).
% We then prompt the LLM to perform a \textbf{Reasoning-Guided Judgment}. For each unit $(p_k, j_k, v_k) \in \mathcal{S}_{\text{sub}}$, we regenerate the judgment conditioned on $z$:
% \begin{equation*}
% \tilde{j}_k = \mathcal{T}_{\text{refine}}(p_k \mid z_{reason}).
% \end{equation*}
% Finally, to explicitly expose this trace, we inject $z_{reason}$ directly into the first unit $\tilde{j}_1$.
% The Final $C_{depth}$ is constructed by serializing the modified units $\{ (p_k, \tilde{j}_k, v_k) \}$.
\paragraph{\textsc{D-CoT} Synthesis.}
We define \textsc{D-CoT} as the expansion of judgment to ensure substantive reasoning soundness by mitigating superficial shortcuts. In contrast to subjective preferences, a ``good'' response in objective correctness tasks depends on rigorous logical constraints (\eg, mathematical proofs or functional code). Normal rationales often fixate on surface-level fluency (\eg, professional tone or formatting) while failing to verify the underlying logical validity.
By enforcing the sequential verification of logical dependencies, sequential thinking provides a deductive rigor that naturally aligns with the strict requirements of objective correctness.
To simulate sequential thinking, we first elicit a Reasoning Trace $z$—a self-solving pass derived from $x$ that explicitly outlines the optimal solution paths required for a correct response. Recognizing that depth-oriented reasoning demands higher cognitive load per unit, we intentionally trade off horizontal coverage for deductive rigor by sampling a focused subset $\mathcal{S}_{\text{sub}} \subset \mathcal{S}$ (typically $|K| \le 3$). In this stage of \textbf{Reasoning-Guided Judgment}, each unit's judgment is re-generated as a derivative of the trace $z$:
\begin{equation*}
\tilde{j}_k = \mathcal{T}{\text{refine}}(p_k \mid z)
\end{equation*}
To ensure the evaluative process is transparent and explicitly grounded in the model's own logic, we inject $z$ directly into the lead unit $\tilde{j}_1$. The final $C_{\text{D}}$ is constructed by serializing these refined units, transforming the verdict into a substantive analytical process anchored by the trace $z$.

\subsection{Mechanism-Adaptive Alignment}
\label{sec:training}
Training proceeds in two stages (Figure~\ref{fig:pipeline}, Panel III): SFT on mixture CoT datasets, followed by GRPO~\citep{shao2024deepseekmath} to align verdicts with human labels.

\paragraph{SFT.}
Following~\citet{frick2025how}, we categorize general RM training data into two domains: \textbf{Preference} (subjective) and \textbf{Correctness} (objective). We construct the mixture dataset $\mathcal{D}_{\text{mix}}$ by assigning $C_{\text{B}}$ to preference instances and $C_{\text{D}}$ to correctness instances.
We first initialize the policy $\pi_\theta$ via SFT on $\mathcal{D}_{mix}$.
Given the $I$, the model is trained to generate the corresponding CoT $c \in \{c_{\text{B}}, c_{\text{D}}\}$ alongside the verdict $v$.
%However, SFT alone relies on imitation and may not fully internalize the optimal mapping between structure and evaluation accuracy.

\paragraph{RLVR via GRPO.}
To optimize verdict accuracy, we employ RLVR via GRPO~\citep{shao2024deepseekmath}, rewarding the model solely for consistency with ground-truth labels:
\begin{equation*}
\begin{split}
\mathcal{J}_{\text{GRPO}}(\theta) = \mathbb{E}_{\substack{I \sim \mathcal{D} \\ \{o_i\} \sim \pi_{\theta_{\text{old}}}}} \Bigg[ & \frac{1}{G} \sum_{i=1}^G \bigg( \frac{\pi_\theta(o_i|I)}{\pi_{\theta_{\text{old}}}(o_i|I)} \hat{A}_i  - \beta \mathbb{D}_{\text{KL}}(\pi_\theta || \pi_{\text{ref}}) \bigg) \Bigg]
\end{split}
\end{equation*}
%where $\hat{A}_i$ is the normalized advantage.
The reward is defined by verdict consistency: a positive reward is assigned if the generated verdict $v_i$ matches the ground-truth human label, and $-1$ otherwise. This process acts as a \textbf{switching amplifier}, inducing an emergent polarization: the model spontaneously learns to couple \textsc{B-CoT} with preference tasks and \textsc{D-CoT} with correctness tasks to maximize rewards, as empirically verified in \S \ref{sec:analysis}. This confirms that the model autonomously converges on the optimal thinking style for each domain.

% Crucially, GRPO enables the model to explore distinct reasoning paths.
% Rather than enforcing rigid rules, this process acts as a switching amplifier, driving the model to autonomously converge on the mechanism ($c_{\text{breadth}}$ vs. $c_{\text{depth}}$) that statistically maximizes verdict accuracy for each task type.
% Our posterior analysis reveals that this results in an emergent Mechanism-Adaptive Plorization, where the policy model spontaneously couples \bcot{ }with preference tasks and \dcot{ }with correctness tasks, as empirically verified in \S \ref{sec:analysis}.

\section{Experiment}
\label{sec:experiment}

%We evaluate the proposed Mix-GRM, focusing on 3 primary objectives:
%(1) \textbf{Overall Performance}: assessing our proposed RM against SoTA baselines across five benchmarks;
%(2) \textbf{Mechanism Performance}: quantifying the uneven benefits of \textsc{B-} versus \dcot{ }at a granular domain level; and
%(3) \textbf{Performance in Downstream Tasks}: validating the reliability of our reward signals in downstream test-time scaling and offline RL applications.

We evaluate Mix-GRM across three objectives: (1) \textbf{Overall Performance} against SoTA baselines; (2) \textbf{Mechanism Efficiency} to quantify the domain-specific benefits of \textsc{B-} and \dcot; and (3) \textbf{Downstream Utility} in Offline RL and Test-time Scaling.

\begin{table*}[t!]
\begin{center}
% 使用 \small 字体而不是 resizebox，保证线条粗细正常
\small
\setlength{\tabcolsep}{3.5pt} % 微调列间距
\begin{tabular}{l c c c c c c c c} 
\toprule

% Header 稍微增加垂直间距
\multirow{2}{*}{\textbf{Models}} & \multirow{2}{*}{\textbf{Stage}} & \multirow{2}{*}{\textbf{Data}} & \multicolumn{6}{c}{\textbf{Benchmarks}} \\
\cmidrule(lr){4-9}
& & & \textsc{RB-v1}$^\dagger$ & \textsc{RB-v2}$^\dagger$ & \textsc{RM-Bench} & \textsc{RMB} & \textsc{PPE} & \textbf{Avg.} \\
\midrule

% --- Reference Group ---
\multicolumn{9}{l}{\textit{Reference: Proprietary Models}} \\
\rowcolor{white}
{\color{gray}~~DeepSeek-V3.2} & {\color{gray}--} & {\color{gray}--} & {\color{gray}95.5} & {\color{gray}92.1} & {\color{gray}91.4} & {\color{gray}83.9} & {\color{gray}69.0} & {\color{gray}86.4} \\
{\color{gray}~~Gemini-3-Flash} & {\color{gray}--} & {\color{gray}--} & {\color{gray}95.3} & {\color{gray}91.1} & {\color{gray}93.8} & {\color{gray}79.2} & {\color{gray}76.4} & {\color{gray}87.2} \\

%\addlinespace[0.5em] % 增加一点间距
\midrule
% --- Open Source Group ---
\multicolumn{9}{l}{\textit{\textbf{Open-Source Baselines}}} \\
~~Skywork-Reward-8B & \badge[red!10]{\textsc{BT}} & \badge[red!10]{44K} & \textbf{93.9} & \textbf{79.7} & 72.4 & 74.4 & 61.7 & 76.5\\
~~JudgeLRM-7B & \badge[blue!10]{\textsc{RL}} & \badge[blue!10]{100K} & 79.0 & 55.6 & 78.5 & 73.1 & 57.9 & 68.8\\
~~RM-R1-7B (Distill) & \badge[green!10]{\textsc{SFT}},\badge[blue!10]{\textsc{RL}} & \badge[green!10]{9K},\badge[blue!10]{64K} & 83.5 & 48.7 & 76.6 & 65.1 & 62.0 & 67.2\\
~~RM-R1-7B (Instruct) & \badge[green!10]{\textsc{SFT}},\badge[blue!10]{\textsc{RL}} & \badge[green!10]{9K},\badge[blue!10]{64K} & 82.3 & 61.4 & 75.1 & 69.9 & 62.0 & 70.1\\
~~FARE-8B & \badge[green!10]{\textsc{SFT}} & \badge[green!10]{2.5M} & 86.3 & 73.4 & 74.1 & \textbf{83.2} & 62.5 & 75.9 \\
~~RubricRM-8B & \badge[green!10]{\textsc{SFT}} & \badge[green!10]{36K} & 86.7 & 71.9 & 74.0 & 78.5 & 62.5 & 74.7\\
~~DeepSeek-GRM-16B & \badge[green!10]{\textsc{SFT}},\badge[blue!10]{\textsc{RL}} & \badge[green!10]{1.2M},\badge[blue!10]{237K} & 76.8 & 56.0 & 63.5 & 70.8 & 59.1 & 65.2\\

%\addlinespace[0.5em] % 增加一点间距
\midrule

% --- Ours Group ---
\multicolumn{9}{l}{\textit{\textbf{Ours: Mix-GRMs}}} \\
% Sub-group SFT
\rowcolor{green!10}
\multicolumn{9}{l}{\footnotesize \hspace{1em}\textit{Stage I: SFT-trained}} \\
~~Base-GRM & \badge[green!10]{\textsc{SFT}} & \badge[green!10]{9K} & 84.5 & 64.7 & 77.0 & 79.2 & 61.1 & 73.3 \\
\rowcolor{gray!8}
~~\textbf{Mix-GRM (Ours)} & \badge[green!10]{\textsc{SFT}} & \badge[green!10]{9K} & 87.2 & 67.8 & \underline{79.2} & 78.9 & 62.1 & 75.1\\
% Sub-group RLVR
\rowcolor{blue!10}
\multicolumn{9}{l}{\footnotesize \hspace{1em}\textit{Stage II: RLVR-trained}} \\
~~Base-GRM & \badge[green!10]{\textsc{SFT}},\badge[blue!10]{\textsc{RL}} & \badge[green!10]{9K},\badge[blue!10]{21K} & 89.0 & 74.0 & 78.8 & 78.5 & \underline{64.0} & \underline{76.9} \\
\rowcolor{gray!8}
~~\textbf{Mix-GRM (Ours)} & \badge[green!10]{\textsc{SFT}},\badge[blue!10]{\textsc{RL}} & \badge[green!10]{9K},\badge[blue!10]{21K} & \underline{91.8} & \underline{77.5} & \textbf{82.7} & \underline{80.1} & \textbf{64.8} & \textbf{79.4}\\

\bottomrule
\end{tabular}
\end{center}
\caption{Performance of RMs on reward benchmarks. Among open-source models, the highest score per column is \textbf{bolded}, and the second-highest is \underline{underlined}. ``Overall'' denotes the average score within each benchmark. Proprietary LLMs (gray rows) are included for reference. $^\dagger$\textsc{RB-v1}/\textsc{v2} refers to RewardBench v1 and v2.}
\label{tab:reward_bench_performance}

\vspace{-1em} % 减少 table 和下方文字的间距
\end{table*}

\subsection{Experimental Setup}
\label{subsec:experimental_setup}

\paragraph{General Reward Benchmarks.} 
We employ five widely recognized benchmarks tailored for general-purpose reward modeling: RewardBench~\citep{lambert2024rewardbench}, RewardBench-v2~\citep{malik2025rewardbench2}, RMB~\citep{zhou2025rmb}, RM-Bench~\citep{liu2025rmbench}, and PPE~\citep{frick2025how}. 
These benchmarks encompass a broad spectrum of tasks, ranging from common tasks like math, coding, and open-ended chat, to specialized capabilities including factuality and instruction-following. 
For Overall Performance, we report standard benchmark-level pairwise comparison accuracy to assess the rewarding capability. 
For granular Mechanism Efficiency analysis, we aggregate instances from these benchmarks and re-categorize them into two fundamental domains, Correctness and Preference, based on their original task metadata. 
Detailed statistics and specific domain mappings for these benchmarks are provided in Appendix~\ref{sec:training_data_synthesis}.

\paragraph{Base Model and Training Data Source.}
We employ Qwen3-8B-Base~\citep{yang2025qwen3technicalreport} trained on a composite corpus $30,000$ samples ($9\text{K}$ SFT, $21\text{K}$ RLVR) spanning five datasets: 
\texttt{HelpSteer3}~\citep{wang2025helpsteer3} (chat, stem \& multilingual), \texttt{Code-Preference} (coding), \texttt{Math-DPO} (math), \texttt{WildGuard}~\citep{han2024wildguard} (safety), and \texttt{OffsetBias}~\citep{park2024offsetbias} (instruction following). 
Detailed sampling protocols and statistical distributions are provided in Appendix~\ref{subsec:data_details}. Other Training Implementation Setting is in Sec.~\ref{sec:training_implementation}.

\paragraph{Baselines.}
We compare our proposed RM with 7 top-tier RMs across two paradigms: 
(1) \textit{Discriminative}: represented by \textbf{Skywork-Reward-v0.2-Llama-3.1-8B}~\citep{liu2024skyworkreward}, a leading scalar model trained via Bradley-Terry modeling; and 
(2) \textit{Generative}: encompassing RL-driven reasoning models (\textbf{JudgeLRM-7B}~\citep{chen2025judgelrmlargereasoningmodels}, \textbf{RM-R1-Instruct}~\citep{chen2025rmr1rewardmodelingreasoning}, \textbf{RM-R1-Distill}, \textbf{DeepSeek-GRM-16B})~\citep{liu2025inference}, synthetic scaling methods (\textbf{FARE-8B}~\citep{xu2025foundational}), and rubric-based approaches \textbf{RubricRM-8B}~\citep{liu2025openrubricsscalablesyntheticrubric}. Notably, RubricRM-8B incorporates two-stage LLMs consisting of a rubric generator and a rubric-based judge.

\begin{table*}[t!]
\begin{center}
\renewcommand{\arraystretch}{1.15} 
\setlength{\tabcolsep}{2.5pt} 
\small

% --- FIXED COMMAND DEFINITIONS (Robust) ---
% \ensuremath ensures these work in both text and math mode automatically
\newcommand{\diffp}[1]{\ensuremath{_{\color{teal}\uparrow #1}}}
\newcommand{\diffn}[1]{\ensuremath{_{\color{magenta}\downarrow #1}}}
\newcommand{\base}[1]{\ensuremath{#1\phantom{_{\uparrow 0.0}}}}
\resizebox{.98\textwidth}{!}{%
\begin{tabular}{l cccccc cccccc c}
\toprule

% --- Headers ---
\multirow{2}{*}{\textbf{Models}} &
\multicolumn{6}{c}{\textbf{Preference Domain}} &
\multicolumn{6}{c}{\textbf{Correctness Domain}} &
\multirow{2}{*}{\textbf{Overall}} \\
\cmidrule(lr){2-7} \cmidrule(lr){8-13}

& \textsc{RB-v1} & \textsc{RB-v2} & \textsc{RM-B}$^\dagger$ & \textsc{RMB} &  \textsc{PPE} & \base{\textbf{Avg.}}
& \textsc{RB-v1} & \textsc{RB-v2} & \textsc{RM-B}$^\dagger$ & \textsc{RMB} &  \textsc{PPE} & \base{\textbf{Avg.}}
& \\
\midrule

% --- Baselines ---
\multicolumn{14}{l}{\textit{\textbf{Baselines}}} \\
~~FARE-8B & 85.0 & 57.3 & 66.9 & \textbf{82.9} & 59.6 & \base{70.4} & 85.2 & 67.3 & 63.0 & 88.1 & 63.3 & \base{73.3} & 71.9 \\
~~RubricRM-8B & 82.4 & 56.0 & 62.2 & 77.5 & \textbf{64.9} & \base{68.6} & 87.6 & 64.2 & 57.6 & 86.5 & 60.4 & \base{71.3} & 70.0\\
~~DeepSeek-GRM & 80.6 & \underline{59.6} & 64.0 & 76.8 & 59.8 & \base{68.2} & 76.6 & 55.8 & 56.6 & 86.8 & 56.8 & \base{66.5} & 67.4\\

\addlinespace[0.3em]
\midrule

% --- Ours Group ---
\multicolumn{14}{l}{\textit{\textbf{Ours: Mix-GRMs}}} \\

% SFT Sub-group
\rowcolor{green!10}
\multicolumn{14}{l}{\footnotesize \hspace{1em}\textit{Stage I: SFT-trained}} \\
% \base command now handles math mode automatically
~~Base-GRM & 81.6 & 55.5 & 63.3 & \underline{80.5} & 60.1 & \base{68.2} & 84.1 & 63.7 & 67.7 & 86.4 & 59.1 & \base{72.2} & 70.2\\
~~Mix-GRM (Breadth) & 83.7 & 59.1 & 65.9 & 77.9 & 59.5 & 69.3\diffp{1.1} & 81.1 & 60.2 & 64.1 & 86.8 & 58.7 & 70.2\diffn{2.0} & 69.8 \\
~~Mix-GRM (Depth) & 80.3 & 50.2 & 70.6 & 70.1 & 58.6 & 65.9\diffn{2.3} & 88.0 & 63.7 & 66.7 & 81.1 & 64.7 & 72.8\diffp{0.6} & 69.4\\
\rowcolor{gray!8}
~~Mix-GRM & 84.9 & 55.7 & 71.2 & 78.7 & 59.2 & 70.0\diffp{1.8} & 88.4 & 65.8 & 67.7 & 81.9 & 63.7 & 73.5\diffp{1.3} & 71.8\\

% RLVR Sub-group
\rowcolor{blue!10}
\multicolumn{14}{l}{\footnotesize \hspace{1em}\textit{Stage II: RLVR-trained}} \\
% Comparison vs. SFT Vanilla (Pref 68.2, Corr 72.2)
~~Base-GRM & 83.0 & 58.0 & 68.5 & 73.8 & 61.4 & 68.9\diffp{0.7} & 89.8 & 69.5 & 69.9 & \textbf{89.5} & 63.4 & 76.4\diffp{4.2} & 72.7 \\
~~Mix-GRM (Breadth) & \textbf{86.2} & 58.8 & 70.1 & 79.2 & 60.7 & \underline{71.0}\diffp{2.8} & 82.8 & 63.4 & 64.3 & 86.5 & 60.7 & 71.5\diffn{0.7} & 71.3 \\
~~Mix-GRM (Depth) & 85.2 & 57.8 & \textbf{75.6} & 75.4 & 61.2 & \textbf{71.0}\diffp{2.8} & \underline{91.8} & \underline{70.3} & \underline{72.9} & 87.4 & \textbf{66.2} & \underline{77.7}\diffp{5.5} & \underline{74.4}\\
\rowcolor{gray!8}
~~\textbf{Mix-GRM} & \textbf{86.2} & \textbf{64.4} & \underline{72.7} & 78.1 & \underline{61.7} & \textbf{72.6}\diffp{3.7} & \textbf{92.2} & \textbf{72.5} & \textbf{74.5} & \underline{88.9} & \underline{65.4} & \textbf{78.7}\diffp{6.5} & \textbf{75.7}\\

\bottomrule
\end{tabular}
}
\end{center}
\caption{Performance of RMs grouped by domain. ``Avg.'' denotes the domain average. We annotate the performance gap relative to the \textit{Base-GRM in SFT} baseline within the same stage using colored subscripts (\textcolor{teal}{$\uparrow$} for gain, \textcolor{magenta}{$\downarrow$} for drop). Highest score per column is \textbf{bolded}, second-highest is \underline{underlined}. $^\dagger$\textsc{RM-B} refers to RM-Bench.}
\label{tab:preference_correctness_performance}
\vspace{-1em}
\end{table*}

\begin{table*}[t!]
\begin{center}
\small
\setlength{\tabcolsep}{4.5pt} % Balanced column spacing
\begin{tabular}{l ccc ccccc}
\toprule

% --- Top Header ---
\multirow{2}{*}{\textbf{Models}} &
\multicolumn{3}{c}{\textbf{Instruction-Following}} &
\multicolumn{5}{c}{\textbf{Mathematical Reasoning}} \\
\cmidrule(lr){2-4} \cmidrule(lr){5-9}

% --- Second Header ---
& \textsc{Alpaca-v2} & \textsc{Arena-Hard} & \textbf{Avg.}
& \textsc{GSM8k} & \textsc{MATH} & \textsc{STEM} & \textsc{TabMWP} & \textbf{Avg.} \\
\midrule

\textbf{SFT} & \text{6.4} & \text{4.2} & \text{5.3} & 75.1 & 25.2 & 38.6 & 40.9 & 45.0 \\

%\addlinespace[0.5em] % 增加一点间距
\midrule

% --- DPO Group ---
\multicolumn{9}{l}{\textit{\textbf{DPO Training (Different RMs)}}} \\
~~$\hookrightarrow$ RubricRM-8B & \text{8.5} & \text{12.5} & \text{10.5} & 76.0 & \underline{26.9} & \textbf{41.4} & 38.8 & \underline{45.9} \\
~~$\hookrightarrow$ FARE-8B & \underline{8.9} & \textbf{15.1} & \underline{12.0} & 75.7 & \underline{26.9} & 39.0 & 41.4 & 45.8 \\
~~$\hookrightarrow$ RM-R1-Instruct & 7.9 & 14.3 & 11.1 & \underline{76.3} & 26.5 & 38.5 & \underline{41.7} & 45.8 \\
~~$\hookrightarrow$ DeepSeek-GRM-16B & 8.0 & 14.1 & 11.1 & 75.6 & 26.6 & 38.7 & 41.6 & 45.6 \\

% --- Ours Row ---
\rowcolor{blue!10}
~~$\hookrightarrow$ \textbf{Ours (Mix-GRM)} & \textbf{9.2} & \underline{15.0} & \textbf{12.1} & \textbf{77.6} & \textbf{27.1} & \underline{39.0} & \textbf{41.9} & \textbf{46.4} \\

\bottomrule
\end{tabular}
\end{center}
\caption{Performance of DPO-trained policy models using different reward models on instruction-following and math-reasoning benchmarks. ``Avg.'' is the average score of all benchmarks in each domain. In each column, the highest score is \textbf{bolded} and the second-highest is \underline{underlined}.}
\label{tab:dpo_performance}
\vspace{-1em}
\end{table*}

\begin{figure}[!t]
\centering
\includegraphics[width=\textwidth]{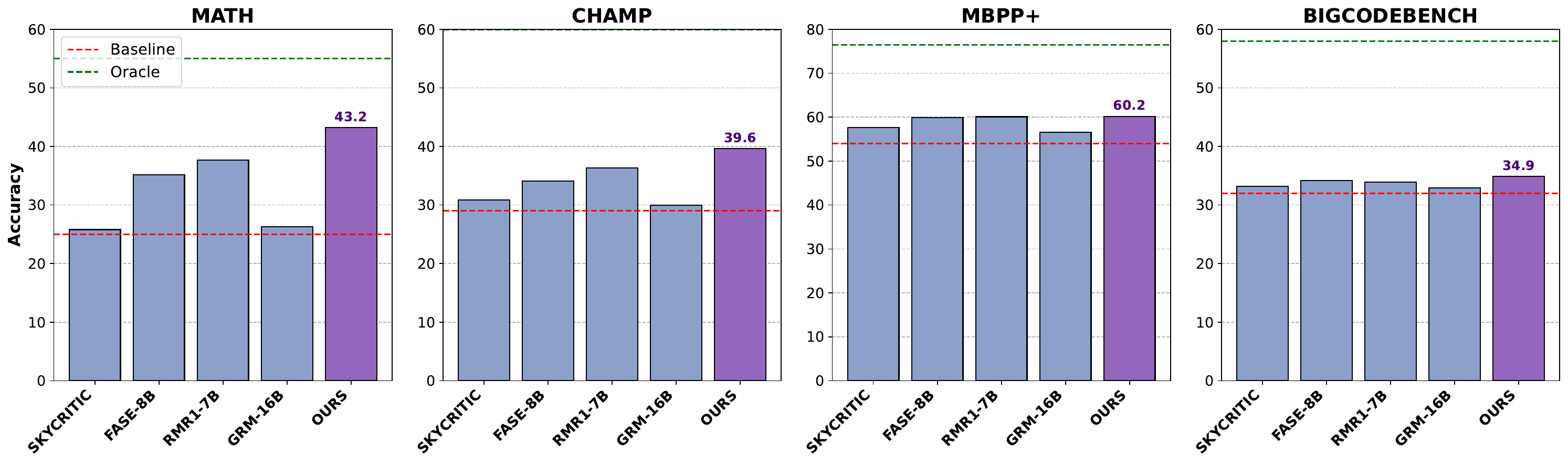}
\caption{
Best-of-$10$ performance across four challenging reasoning and coding benchmarks. Mix-GRM (ours) consistently achieves the highest accuracy across all tasks, effectively identifying solutions in both mathematical and code generation scenarios. \textcolor{red}{Red} and \textcolor{green}{green} lines denote random and oracle selection baselines.
}
\label{fig:response_reranking}
\vspace{-1em}
\end{figure}
\subsection{Overall Performance in Benchmarks}
\label{subsec:overall_performance}
Table~\ref{tab:reward_bench_performance} validates the effectiveness of our Mix-GRM through three dimensions. 
\paragraph{Effectiveness of Mixture SFT}: Via mixture SFT alone, \textit{Mix-GRM} achieves a remarkable average score of $75.1$. This performance surpasses GRMs requiring computationally expensive RL to elicit long-CoT capabilities—outperforming \textit{RM-R1-Instruct} by $5.0$ and \textit{DeepSeek-GRM-16B} by $9.9$. Furthermore, it beats \textit{RubricRM-8B} ($+0.4$), which relies on a complex but static rubric-template CoT. This confirms that aligning reasoning mechanisms serves as a potent alternative strategy, alongside approaches focused on RL exploration or static template engineering.
\paragraph{Superiority of Data Efficiency}: Mix-GRM achieves these gains with substantially less data. While \textit{FARE-8B} relies on massive scaling ($\approx 2.5\mathrm{M}$ samples) to reach $75.9$, \textit{Mix-GRM} attains a comparable $75.1$ in the SFT stage using merely $9\mathrm{K}$ samples. This finding highlights that optimizing CoT mechanisms yields a substantially higher training signal density, enabling data efficiency compared to brute-force dataset expansion.
\paragraph{Switching Amplification via RLVR}: Mix CoT maximizes the efficacy of the RLVR stage, unlocking greater performance gains than unstructured CoT. RLVR boosts \textit{Mix-GRM} by $4.3$ ($75.1 \rightarrow 79.4$), compared to a $3.6$ gain for \textit{Base-GRM} ($73.3 \rightarrow 76.9$). Consequently, the performance gap over the \textit{Base-GRM} widens from $1.8$ (SFT) to $2.5$ (RLVR), confirming that the aligned mechanism offers a more exploitable base for the RL. Furthermore, our subsequent analysis (Sec~\ref{sec:analysis}) reveals that these gains are fundamentally underpinned by an emergent polarization in mechanism allocation, where RLVR sharpens the model's reasoning style to match task-specific demands.

\subsection{Mechanism Efficiency}
\label{subsec:mechanism_performance}
Table~\ref{tab:preference_correctness_performance} reveals that mechanism efficacy is strictly task-dependent. In the SFT stage, we observe a \textbf{distinct performance trade-off}: \bcot{} improves Preference via lateral coverage but degrades Correctness ($72.2 \rightarrow 70.2$), whereas \dcot{} enhances deductive soundness but fails in Preference ($68.2 \rightarrow 65.9$). These results indicate that simply extending CoT length does not guarantee universal gains; while principle expansion facilitates multi-dimensional evaluation, it offers no inherent advantage for deep reasoning. However, \textit{Mix-GRM} overcomes these limitations through a \textbf{synergistic mutual enhancement}. By integrating orthogonal strengths, it not only surpasses the \textit{Base-GRM} ($70.2 \rightarrow 71.8$) but surprisingly outperforms specialized single-mode models on their respective strongholds (\eg, exceeding Depth-only on Correctness). This synergy becomes critical during the RLVR stage, where single-mode mechanisms encounter hard performance ceilings—most notably, \textit{Mix-GRM (Breadth)} plateaus on correctness tasks. In contrast, the \textit{Mix-GRM} enables RL optimization to reach a superior ceiling ($78.7$). This confirms that the \textbf{CoT structure itself acts as a bottleneck for RL optimization}; the mixed structures do not merely inherit component strengths but construct a robust reasoning framework that transcends the inherent limitations of isolated mechanisms.

\subsection{Downstream Utility}
\label{subsec:performance_downstream_tasks}

To validate the practical utility of \textit{Mix-GRM}, we apply it to two downstream applications: (i) serving as a reward signal for \textbf{Offline Reinforcement Learning}, and (ii) acting as a verifier for \textbf{Test-time Scaling}. We provide detailed descriptions of these application settings in Appendix~\ref{sec:evaluation_implementation}.
\paragraph{Reward Model for Offline Reinforcement Learning.}
In Offline RL via Direct Preference Optimization (DPO)~\citep{rafailov2023direct}, RM constructs high-quality preference pairs $(y_w, y_l)$ to supervise policy alignment.
Table~\ref{tab:dpo_performance} shows that models trained on these signals achieve a peak win rate of \textbf{12.1} in instruction-following, surpassing \textit{FARE-8B} (12.0) and \textit{RubricRM} (10.5).
Crucially, this alignment gain does not compromise reasoning capabilities; in the math domain, Mix-GRM maintains a SOTA accuracy of \textbf{46.4}, edging out \textit{RubricRM} (45.9) and \textit{RM-R1-Instruct} (45.8).
Specifically, Mix-GRM achieves $77.6\%$ on GSM8K, demonstrating a clear lead over the SFT baseline ($75.1\%$).
These results confirm that Mix-GRM provides reliable supervision, enabling policies to internalize both helpfulness and correctness.

\paragraph{Reward Model for Test-time Scaling.}
% \begin{wrapfigure}{r}{0.55\textwidth}
% \centering
% \includegraphics[width=0.52\textwidth]{figure/response_reranking.pdf} % 也可以将其改为 width=\linewidth
% \caption{
% Best-of-$10$ performance across four challenging reasoning and coding benchmarks. Mix-GRM (ours) consistently achieves the highest accuracy across all tasks, effectively identifying solutions in both mathematical and code generation scenarios. \textcolor{red}{Red} and \textcolor{green}{green} lines denote random and oracle selection baselines.
% }
% \label{fig:response_reranking}
% \vspace{-1em}
% \end{wrapfigure}
For test-time scaling, leveraging increased inference-time compute to enhance generalization, Mix-GRM functions as a robust verifier to re-rank candidates to identify the optimal solution via Best-of-$N$ selection.
Following the JETTS protocol~\citep{zhou2025evaluating}, we evaluate $N=10$ samples from a Llama-3.1-8B generator across 4 diverse benchmarks: MATH and CHAMP (math), as well as MBPP+ and BigCodeBench (coding). As shown in Figure~\ref{fig:response_reranking}, our method consistently secures the highest accuracy, setting a new SOTA for 8B-scale rerankers. The performance advantage is particularly pronounced in reasoning-heavy tasks; for instance, on MATH, our model achieves an accuracy of $43.2\%$, outperforming the RL-driven \textit{RM-R1} ($37.7\%$) and the data-intensive \textit{FARE-8B} ($35.2\%$). This confirms that ours provides a more discriminative signal for logical verification than methods relying on massive data scaling or generic RL.

\begin{wrapfigure}{r}{0.49\textwidth}
\centering
\includegraphics[width=0.48\textwidth]{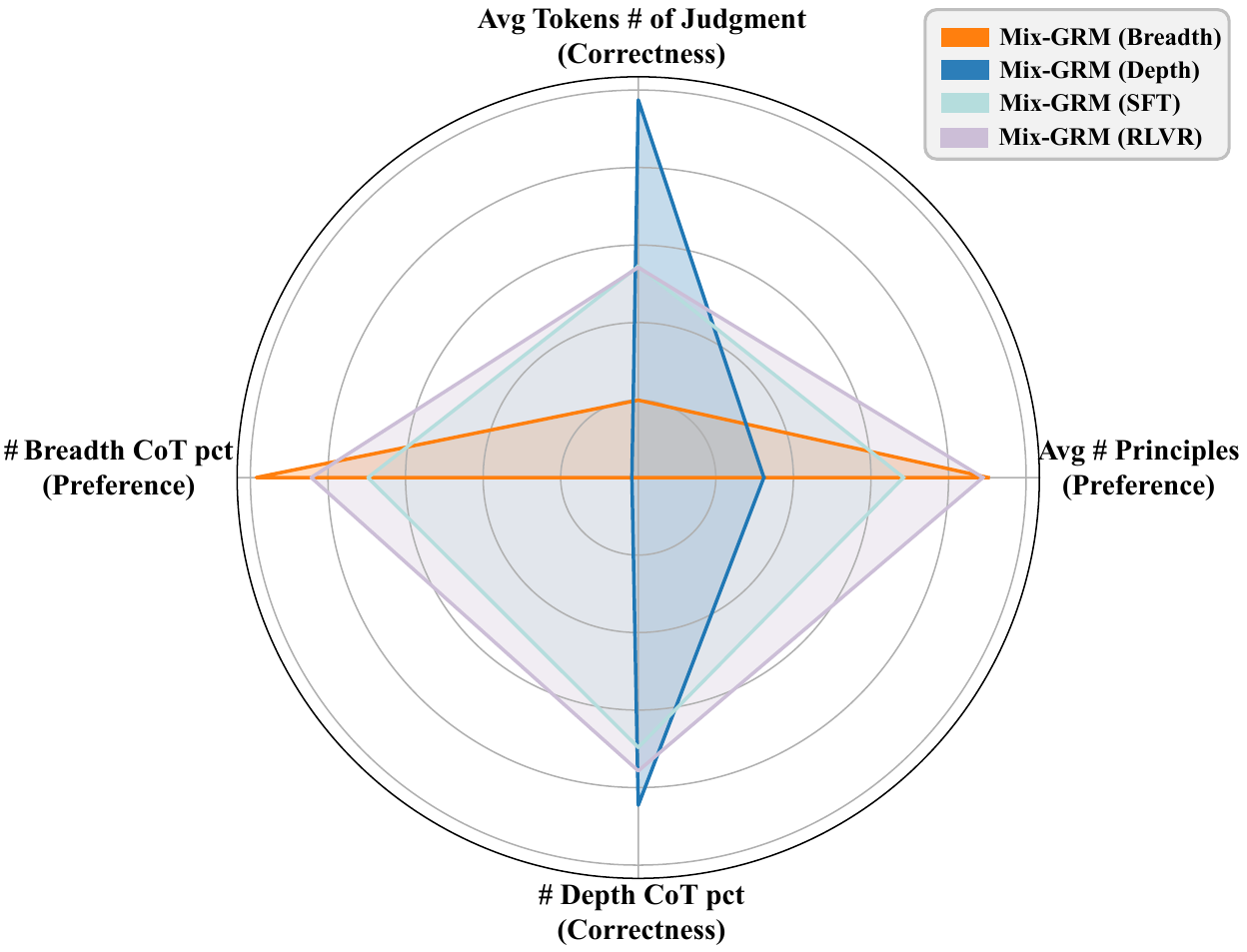}
\caption{
\textbf{Structural evolution of CoT mechanisms.} The chart tracks 4 indicators: the average token length per judgment, average principle count, and the percentage of CoT classified as having Breadth or Depth characteristics. 
}
\label{fig:sft_rlvr_cot}
\vspace{-1em}
\end{wrapfigure}

\section{Analysis}
\label{sec:analysis}
%\paragraph{Morphological Analysis.} We visualize the structural transformation of evaluation CoT to understand how our pipeline reshapes reasoning morphology. As illustrated in Figure \ref{fig:sft_rlvr_cot}, the geometric evolution from single-mode baselines to our post-trained model reveals three phases of transformation. First, the distinct polarization of the Breadth and Depth baselines empirically confirms the structural rigidity of static templates: relying on a single CoT style inevitably creates a capability blind spot, sacrificing either deductive depth or semantic coverage. Second, the emergence of the balanced profile in \textbf{Fuse\dcot (SFT)} indicates that the model has successfully internalized morphological plasticity. Most significantly, the global expansion in the \textbf{RLVR stage} provides empirical validation of our core hypothesis regarding morphological optimality. The resulting profile—simultaneously pushing the boundaries of depth (vertical) and breadth (horizontal)—indicates that the model has spontaneously converged on domain-specific structural biases. Driven by the optimization for verdict accuracy, the model autonomously amplifies deductive depth as the essential strategy for correctness tasks, while reinforcing comprehensive breadth for preference tasks. This emergent specialization confirms that our proposed morphologies are not merely heuristic constraints but the intrinsic optimal structures required to solve their respective alignment challenges.

\paragraph{Switching CoT Mechanism Analysis.} Visualizing structural transformations (Figure \ref{fig:sft_rlvr_cot}) reveals how our pipeline reshapes reasoning mechanisms.
Here, Single-mode strategies show extreme trade-offs: \textbf{Mix-GRM (Breadth)} expands horizontally (high principle count), while \textbf{Mix-GRM (Depth)} extends vertically (long judgments). In contrast, \textbf{Mix-GRM (SFT)} achieves a robust union of both, which is further expanded into a broader reasoning manifold by \textbf{RLVR}.
First, the polarization of Breadth and Depth baselines confirms the rigidity of static templates, which create capability blind spots by sacrificing either reasoning depth or semantic coverage. Second, the balanced profile of \textbf{Mix-GRM (SFT)} indicates successful internalization of different distinct mechanisms. Most pivotally, the global expansion during \textbf{RLVR} validates our hypothesis of mechanism polarization. By optimizing for verdict accuracy, the model spontaneously converges on domain-specific mechanism biases—amplifying \dcot{ }for correctness while reinforcing \bcot{ }for preference. This emergent specialization confirms that our proposed alignment is not a handcrafted heuristic, but an inherent structural necessity discovered by the model to maximize evaluation efficacy.

\paragraph{Emergent Polarization Analysis.} To determine whether the emergent polarization toward domain-specific reasoning (\ie, \bcot{} for Preference and \dcot{} for Correctness) is actively driven by RLVR or merely inherited from SFT priors, we analyzed the distribution of generated CoT structures on the test set using specific structural indicators (\eg, principle counts and trigger phrases). Although the SFT training data was completely domain-aligned, the post-SFT model achieved only a 73\% structural match rate during inference. Remarkably, following RLVR—which relies exclusively on final verdict supervision without explicit structural labels—this alignment surged to 95\%. This substantial improvement confirms that the model transcends static SFT priors, autonomously learning to select and optimize the most effective reasoning structure for each domain during the reinforcement learning phase.
% \begin{figure}[ht]
% \centering
% \includegraphics[width=.48\textwidth]{figure/sft_rlvr_cot.pdf}
% \caption{
% \textbf{Structural evolution of CoT mechanisms.} The chart tracks 4 indicators: the average token length per judgment, average principle count, and the percentage of CoT classified as having Breadth or Depth characteristics. Single-mode strategies show extreme trade-offs: \textbf{Mix-GRM (Breadth)} expands horizontally (high principle count), while \textbf{Mix-GRM (Depth)} extends vertically (long judgments). In contrast, \textbf{Mix-GRM (SFT)} achieves a robust union of both, which is further expanded into a broader reasoning manifold by \textbf{RLVR}.
% }
% \label{fig:sft_rlvr_cot}
% \vspace{-1em}
% \end{figure}

\begin{figure*}
\centering
\includegraphics[width=.8\textwidth]{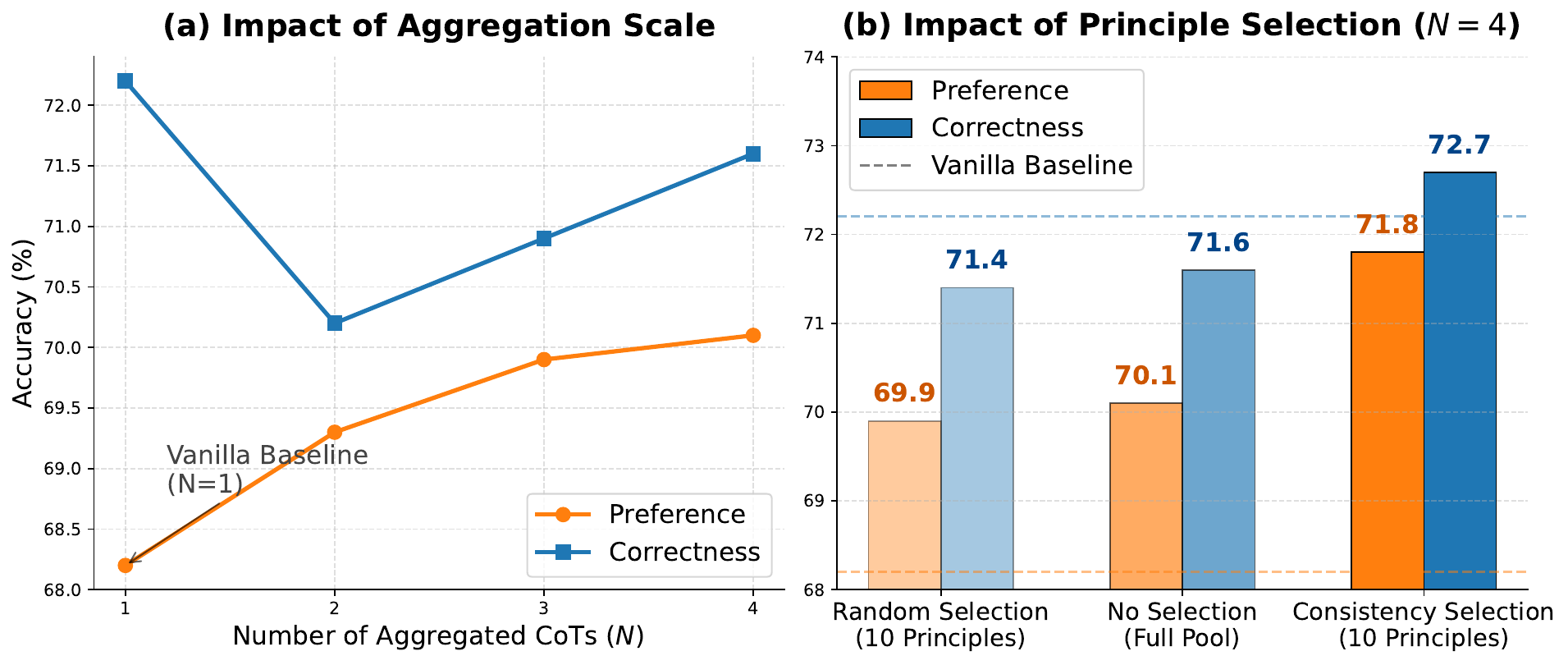}
\caption{
\textbf{Ablation of \bcot{ }synthesis.} (a) Aggregation Scale: Performance as aggregated rationales ($N$) increases from 1 (Vanilla) to 4. (b) Principle Selection: Comparison of Random, Full, and Consistency (Top-10) selection from the $N=4$ pool. \textcolor{orange}{Orange}/\textcolor{blue}{blue} lines denote Preference/Correctness; dashed lines indicate the Vanilla baseline.
}
\label{fig:ablation_study}
\vspace{-1em}
\end{figure*}

\paragraph{Scaling \& Selection Analysis.} To understand the mechanics of \bcot, we decouple the impact of quantity (aggregation scale) from quality (principle selection) as shown in Figure \ref{fig:ablation_study}.1) Quantity (Scaling): Figure \ref{fig:ablation_study}(a) demonstrates that performance improves monotonically as the number of parallel CoTs ($N$) increases from 1 to 4. This confirms that ``breadth" functions by expanding coverage; by aggregating diverse perspectives, the model minimizes the risk of overlooking critical error patterns.2) Quality (Selection): However, more is not always better. Figure \ref{fig:ablation_study}(b) compares three strategies within the $N=4$ pool: $\text{Breadth}_{\text{Rand}}$, $\text{Breadth}_{\text{Full}}$, and $\text{Breadth}_{\text{Top10}}$, where Top-10 means ten most frequent principles appearing across 4 CoTs. We observe a clear hierarchy: $\text{Breadth}_{\text{Top10}} > \text{Breadth}_{\text{Full}} > \text{Breadth}_{\text{Rand}}$. While the full pool improves over random sampling, it is the Top-10 consensus that achieves the highest gains ($71.8/72.7$).
This suggests a denoising effect where low-frequency principles introduce noise, while high-frequency ones form a more robust ``reasoning consensus.'' Thus, representativeness—not just volume—is vital for robust breadth.

\paragraph{Computational Overhead and Token Cost Analysis.} To ensure a transparent comparison of computational overhead, we designed our pipelines to maintain strict algorithmic fairness and token parity across reasoning styles. First, regarding data synthesis, both \bcot{} (which samples twice and merges rationales) and \dcot{} (which generates a solve trace before synthesizing the final CoT) utilize approximately two reasoning passes. Notably, we strictly prohibit introducing new information during the merge phase, limiting it solely to merging and deduplication; therefore, it does not constitute an additional reasoning pass. This parity ensures that observed performance differences stem from structural efficacy rather than raw compute disparities. Second, as detailed in Table \ref{tab:token_cost}, the token consumption during Data Synthesis, RLVR Training, and Inference remains highly comparable across configurations. These marginal differences confirm that \bcot, \dcot, and our adaptive Mix-CoT operate within the exact same order of magnitude of compute cost.

\begin{table*}[h]
    \centering
    \small
    \begin{tabular}{lcccc}
        \toprule
        \textbf{CoT Style} & \textbf{Reasoning Passes} & \textbf{Avg. SFT Tokens} & \textbf{Avg. RLVR Rollouts} & \textbf{Avg. Inference Tokens} \\
        \midrule
        D-CoT & 2 (Trace + Gen) & 624 & 682 & 702 \\
        B-CoT & 2 ($N=2$ Sample) & 711 & 830 & 824 \\
        Mix-CoT (Ours) & Adaptive & 648 & 725 & 731 \\
        \bottomrule
    \end{tabular}
    \caption{Accounting table reporting the compute and token costs. Token consumption across our proposed methods remains within the same order of magnitude.}
    \label{tab:token_cost}
\end{table*}

\paragraph{Case Study.}
%Table \ref{tab:case_study_detail} (Appendix \ref{sec:case_study}) elucidates the structural drivers of the double dissociation. In Case 1 (Preference), \bcot{ }acts as a multi-dimensional scanner, identifying lateral mismatches (\eg, language) that \dcot{ }overlooks due to attentional tunneling. Conversely, Case 2 demonstrates that \dcot{ }provides a derivation, serving as a probe that exposes factual hallucinations (\eg, $K > Mg$) hidden within lengthy responses. Here, \bcot{ }fails due to feature interference, mistaking superficial formatting for logical validity. This confirms that while Breadth ensures multi-faceted alignment, Depth remains the non-negotiable driver for rigorous deductive verification.
Table \ref{tab:case_study_detail} elucidates the structural mechanisms behind the observed double dissociation. Case 1 demonstrates why \bcot{} dominates preference tasks: acting as a multi-dimensional scanner, it successfully penalizes a detailed but language-mismatched response by validating lateral constraints (\eg, \textit{Linguistic Alignment}), whereas \dcot{} exhibits attentional tunneling,'' fixating on verifying historical facts while missing the high-level language mismatch. Conversely, Case 2 reveals why \dcot{} is essential for correctness: its step-by-step derivation acts as a logic probe, allowing it to spot subtle factual hallucinations (\eg, $K > Mg$) hidden within a lengthy explanation. Here, \bcot{} actively fails due to feature interference,'' where it mistakes superficial comprehensiveness (length and formatting) for logical validity. This confirms that while Breadth is necessary for satisfying diverse user preferences, Depth is the non-negotiable driver for rigorous verification.

\begin{table*}[ht!]
\caption{Case Study. \textbf{Case 1} shows how \bcot{} aggregates diverse principles to identify subtle preference nuances. \textbf{Case 2} shows how \dcot{} performs step-by-step verification to catch logical errors.}
\label{tab:case_study_detail}
\centering

\scriptsize 
\renewcommand{\arraystretch}{1.5} % 稍微增加行高，防止拥挤
\setlength{\tabcolsep}{2pt}       % 【优化2】极限压缩列间距

\begin{tabularx}{\textwidth}{@{} p{1.8cm} X @{}}
\toprule

% ================= CASE 1 =================
\multicolumn{2}{c}{\cellcolor{orange!20}\textbf{\textsc{Case 1: Preference Domain (\bcot{} Wins)}}} \\

% --- Context Panel ---
\multicolumn{2}{l}{\cellcolor{headergray}\textit{\textbf{Instruction:}} \jap{アフガニスタンがパキスタンの傀儡というのは本当ですか？} (Is it true that Afghanistan is a puppet of Pakistan?)} \\
\multicolumn{2}{l}{
    % 内部表格继续使用 linewidth
    \begin{tabularx}{\linewidth}{@{} X | X @{}}
        \textbf{Response A (Rejected):} & \textbf{Response B (Chosen):} \\
        \scriptsize \textbf{[Language: English]} "A sensitive topic! ...While Pakistan has historically exerted significant influence..." & 
        \scriptsize \textbf{[Language: Japanese]} \jap{「アフガニスタンとパキスタンの関係については... 簡単に『傀儡』と断言するのは適切ではなく...」}
    \end{tabularx}
} \\
\midrule
% --- Reasoning Panel ---
\multicolumn{2}{l}{\textbf{Reasoning Comparison}} \\

% 1. Vanilla
\xmark~\modelbadgel{Vanilla-CoT} & 
Assistant A offers a comprehensive breakdown of historical context... \textbf{Verdict: [[A]]} (Fail: Ignored language mismatch) \\ \hline

% 2. Breadth
\cmark~\modelbadgeh{orange!30}{\bcot} & 
\cellcolor{breadthorange}
1. \textbf{Principle of Linguistic Alignment:} "Assistant B's response is in Japanese... Sub-Verdict: <<B>>" \newline
2. \textbf{Principle of Contextual Nuance:} "Assistant A provides a detailed explanation... Sub-Verdict: <<A>>" \newline
3. \textbf{Principle of Cultural Sensitivity:} "..." \newline
\textbf{Final Verdict: [[B]]} (Success) \\ \hline

% 3. Depth
\xmark~\modelbadgel{\dcot} & 
1. \textbf{Principle of Comprehensive Contextualization:} The evaluation needs deep reasoning, and let's solve the instruction: \jap{アフガニスタンが...}. Based on this, Assistant A excels... Sub-Verdict: <<A>> \newline
2. \textbf{Principle of Balanced Presentation:} ... \newline
\textbf{Final Verdict: [[A]]} (Fail: Tunnel vision on content depth) \\
\bottomrule
\addlinespace[1.5em] 

% ================= CASE 2 =================
\multicolumn{2}{c}{\cellcolor{blue!15}\textbf{\textsc{Case 2: Correctness Domain (\dcot{} Wins)}}} \\

% --- Context Panel ---
\multicolumn{2}{l}{\cellcolor{headergray}\textit{\textbf{Instruction:}} On the basis of oxidation-reduction potential, which of the following is most likely to occur? ...} \\
\multicolumn{2}{l}{
    \begin{tabularx}{\linewidth}{@{} X | X @{}}
        \textbf{Response A (Chosen):} & \textbf{Response B (Rejected):} \\
        \scriptsize "The reactivity of metals... The order is: Alkali > ... > Zn > ... > Ag. \newline Based on this, the most likely reaction is: \textbf{D. Zn + 2Ag(NO3)...}" (Correct) & 
        \scriptsize "The reactivity... Alkali > Alkaline earth... \newline Analysis: \textbf{B. Mg + 2KNO3 -> 2K + Mg(NO3)2}." (\textbf{Error:} K > Mg.)
    \end{tabularx}
} \\
\midrule

% --- Reasoning Panel ---
\multicolumn{2}{l}{\textbf{Reasoning Comparison}} \\

% 1. Vanilla
\xmark~\modelbadgel{Vanilla-CoT} & 
To evaluate the responses, let's consider factors: 1. Helpfulness; 2. Relevance... Assistant B goes further by analyzing multiple options... \textbf{Final Verdict: [[B]]} (Fail: Fooled by length/detail) \\ \hline

% 2. Breadth
\xmark~\modelbadgel{\bcot} & 
1. \textbf{Principle of Comprehensive Option Analysis:} "Assistant B analyzes all provided options (A-H)... Sub-Verdict: <<B>>" \newline
2. \textbf{Principle of Informative Detail:} "Assistant B includes more elements... Sub-Verdict: <<B>>" \newline
\textbf{Final Verdict: [[B]]} (Fail: Superficial heuristic) \\ \hline

% 3. Depth
\cmark~\modelbadgeh{blue!20}{\dcot} & 
\cellcolor{depthblue}
1. \textbf{Principle of Comprehensive Analysis:} The evaluation needs to deep reasoning, and let's solve the instruction: To determine the most likely reaction... The correct order is Alkali > ... Given the options, the most likely reaction is Option D. ... Assistant B correctly identifies the importance but incorrectly identifies Option B... Sub-Verdict: <<A>> \newline
2. \textbf{Principle of Direct Relevance:} Assistant A directly addresses the question... Sub-Verdict: <<A>> \newline
\textbf{Final Verdict: [[A]]} (Success) \\

\bottomrule
\end{tabularx}
\end{table*}

\section{Conclusion}
\label{sec:conclusion}

%This work characterizes the impact of reasoning scaling in GRMs through the lens of morphology. Our investigation reveals a fundamental double dissociation: Breadth-CoT provides the lateral coverage essential for subjective preference alignment, while Depth-CoT supplies the vertical rigor required for objective correctness. Crucially, we demonstrate that simply increasing CoT length without morphological alignment yields diminishing returns or even performance degradation.

%By introducing the MA-GRM framework, we show that RMs can effectively bridge this gap by adopting task-specific structural biases. Our findings suggest that the next frontier of reward modeling lies not in the brute-force expansion of how much a model writes, but in the precision of how it structures its reasoning process. By transitioning from unstructured lengthening toward adaptive reasoning morphology, we enable the development of more robust and efficient RMs that are intrinsically aligned with the diverse nature of human queries.

This work demonstrates that beyond mere length scaling, the reliability of GRMs is fundamentally driven by the integration of different reasoning mechanisms. 
%We reveal that blindly adhering to a singular reasoning paradigm—whether focused solely on breadth or depth—fails to achieve optimal performance across general tasks and can even lead to performance degradation.
By introducing Mix-GRM, we prove that the frontier of reward modeling lies in synergizing two orthogonal reasoning mechanisms: \textsc{B-CoT} for multi-dimensional coverage and \textsc{D-CoT} for judgment soundness. Through mechanism-adaptive alignment, Mix-GRM ensures that the RM's reasoning mechanism is precisely calibrated to the nature of the task. Ultimately, these findings shift the focus of GRM development from brute-force expansion to structural optimization.

\section*{Limitations}

While Mix-GRM significantly enhances evaluation reliability through mechanism alignment, we identify two primary limitations that warrant further investigation:

\paragraph{Granularity of the Reasoning Manifold.} Our framework successfully captures the double dissociation between Subjective Preference and Objective Correctness, which we identify as the dominant axes of the reasoning manifold. However, this dichotomy represents a coarse-grained mapping of the diverse alignment landscape. Real-world tasks often exist on a continuous spectrum or involve hybrid demands that intricately blend deductive rigor with multi-dimensional nuances. While we prove that the model's reasoning structure spontaneously converges toward these two primary poles, our current categorization may act as a low-rank approximation of a higher-dimensional space of mechanisms. Future work could explore more granular taxonomies to achieve even more precise task-mechanism calibration.

\paragraph{Rigidity in Explicit Hybrid Tasks.}While our analysis demonstrates that RLVR induces an intrinsic convergence toward specialized reasoning poles, this emergent polarization may introduce structural rigidity when encountering highly complex, cross-domain scenarios. The current framework excels at aligning specialized mechanisms with their respective domains, but our findings ultimately prove that ``one size does not fit all'' for reward structures. Emerging real-world applications, such as agentic Deep Research, increasingly demand an explicit, dynamic fusion of rigorous deductive logic and high-quality, stylistic writing. The spontaneous sharpening of reasoning styles might currently come at the cost of this generalist flexibility. Therefore, a logical next step for future research is to develop dedicated hybrid slicing benchmarks to explicitly evaluate the trade-off between style and logic, alongside the design of more sophisticated, fine-grained hybrid structures (e.g., soft-routing mechanisms) that can fluidly transition across the reasoning manifold.

\bibliography{iclr2025_conference}
\bibliographystyle{iclr2025_conference}

\appendix

\newpage

\section{Training Implementation}
\label{sec:training_implementation}

\subsection{Hyperparameters Setting}
\label{sec:hyperparameters}
We provide the detailed hyperparameter settings in the Table~\ref{tab:hyperparameters-sft} and Table~\ref{tab:hyperparameters-rl}.
\begin{table}[!h]
\centering
\begin{minipage}[t]{0.48\textwidth}
\small
\centering
\resizebox{\textwidth}{!}{%
\begin{tabular}{ll}
\toprule
\textbf{Hyperparameters} & \textbf{Values} \\
\midrule
Epochs & $2$ \\
Learning rate & $2\mathrm{e}{-5}$ \\
Batch Size & $128$ (gradient accumulation steps $= 16$) \\ 
Seq Length & $12,288$ \\ 
Weight Decay & $0.$ \\
Warmup & 5\% linear warmup\\
\bottomrule
\end{tabular}%
}
\caption{Hyperparameter settings for SFT.}
\label{tab:hyperparameters-sft}
\end{minipage}
\hfill
\begin{minipage}[t]{0.48\textwidth}
\small
\centering
\resizebox{\textwidth}{!}{%
\begin{tabular}{ll}
\toprule
\textbf{Hyperparameters} & \textbf{Values} \\
\midrule
Training Steps & $100$ \\
Learning Rate & $1\mathrm{e}{-6}$\\
Batch Size & $128$ \\ 
KL Loss Coefficient  & $0.001$ \\
KL Coefficient  & $0.001$ \\
Rollouts & n$=8$ using vLLm with temperature $0.8$\\
\bottomrule
\end{tabular}%
}

\caption{Hyperparameter settings for RL.}
\label{tab:hyperparameters-rl}
\end{minipage}
\end{table}

\subsection{Training Data Source Details}
\label{subsec:data_details}

To cultivate general rewarding capabilities, it is essential to curate a training corpus that encompasses diversified real-world scenarios. 
We construct our dataset by performing stratified random sampling from representative data sources, ensuring balanced coverage across distinct alignment domains, including general chat, STEM, coding, math, safety, multilingual, and instruction following.
The specific source datasets, their corresponding domains, and the sampling statistics are detailed in Table~\ref{tab:data_stats}.

\begin{table}[ht]
\centering
\vspace{2mm} % Add a little space
\resizebox{0.48\textwidth}{!}{
\begin{tabular}{llc}
\toprule
\textbf{Source Dataset} & \textbf{Domain} & \textbf{Samples} \\
\midrule
\multirow{4}{*}{\textbf{\texttt{HelpSteer-3}} (Single-Turn)} & General Chat & 4,973 \\
                                            & STEM & 2,321 \\
                                            & Code & 4,322 \\
                                            & Multilingual & 3,260 \\
\cmidrule{1-3}
\textbf{\texttt{Code-Preference}} & Code & 4,000 \\
\textbf{\texttt{Math-DPO}} & Math & 4,000 \\
\textbf{\texttt{WildGuard}} & Safety & 4,000 \\
\textbf{\texttt{OffsetBias}} & Instruction Following & 4,000 \\
\midrule
\textbf{Total} & -- & \textbf{30,876} \\
\bottomrule
\end{tabular}}
\caption{Composition and statistics of the training data sampled from domain-specific sources.}
\label{tab:data_stats}
\end{table}

\subsection{Training Data Synthesis Details}
\label{sec:training_data_synthesis}
To synthesize the CoT data for SFT, we utilized \texttt{DeepSeek-v3} (0324 snapshot) as the backbone generator. 
The generation process was configured with a sampling temperature of $T=0.8$ to promote diversity in the trajectories while maintaining logical coherence.
Notably, we abstain from consistency filtering:
Contrary to common practices that discard samples where the synthesized verdict diverges from the ground-truth human label, our empirical verification reveals that training on the full synthesized CoTs yields superior performance compared to aggressive filtering, regardless of verdict consistency.

\subsection{Training Offline Reinforcement Learning Details}

To strictly control for temporal data leakage and ensure a fair comparison with the release dates of our evaluation benchmarks, we select \textbf{Llama-3-8B} as our base foundation model. The offline reinforcement learning pipeline consists of two phases: SFT initialization and DPO.

\paragraph{Policy Initialization (SFT).}
We first derive a supervised policy model by fine-tuning Llama-3-8B on a composite dataset. This dataset ensures basic instruction-following and reasoning capabilities, consisting of the \textbf{UltraChat} dataset \citep{ding2023enhancing} and a random subset of $40\mathrm{K}$ samples from \textbf{MetaMathQA} \citep{yu2024metamath}.
We train the model for $2$ epochs using a learning rate of $2\mathrm{e}{-5}$ and a maximum sequence length of $2,048$ tokens. This SFT model serves as the initial policy $\pi_{\text{ref}}$ for the subsequent DPO stage.

\paragraph{DPO Data Construction via RM Labeling.}
To evaluate the practical utility of different RMs, we employ them to annotate preferences on a unified source dataset.
The prompt source comprises $10\mathrm{K}$ instructions randomly sampled from \textbf{UltraFeedback} \citep{cui2024ultrafeedback} and $40\mathrm{K}$ instructions from \textbf{MetaMathQA}.
For each instruction $x$, we generate $N=5$ diverse candidate responses using \texttt{gpt-4o-mini} with a temperature of $0.8$.

We adopt a \textbf{Pairwise Scoring Aggregation} strategy to construct the final preference pairs $(x, y_w, y_l)$. Specifically, for the set of 5 responses, we generate all possible combinations of pairs ($\binom{5}{2} = 10$ pairs). The target RM evaluates each pair, assigning $+1$ point to the preferred response (chosen) and $0$ to the non-preferred one (rejected). After traversing all pairs, we calculate the cumulative score for each response. The response with the highest total score is selected as the positive sample ($y_w$), and the response with the lowest total score is selected as the negative sample ($y_l$). These labeled pairs are then used to train the policy via DPO.

\section{Evaluation Implementation}
\label{sec:evaluation_implementation}
\subsection{Core Benchmarks}
\label{sec:core_benchmarks}
There is a list of benchmarks and corresponding task coverage.

\begin{table}[t]
\centering
\caption{Task coverage of the evaluated general reward benchmarks.}
\label{tab:benchmark_coverage}
\vspace{2mm}
% Resizing to \linewidth is safer than \textwidth in two-column formats
\resizebox{.48\textwidth}{!}{
\begin{tabular}{lcc} 
\toprule
\textbf{Benchmark} & \textbf{Tasks} & \textbf{Samples}\\
\midrule
\textbf{\textsc{RewardBench}}    & Chat, Math, Code, Safety & 2,985 \\
\textbf{\textsc{RewardBench-v2}} & Focus, IF, Factuality, Math, Safety, Ties & 1,865\\
\textbf{\textsc{RM-Bench}}       & Chat, Math, Code, Safety & 11,943\\
\textbf{\textsc{RMB}}            & Harmfulness, Helpfulness (General, Code) & 14,725 \\
\textbf{\textsc{PPE} }(Exclude Tie)           & Chat, MMLU-Pro, GPQA, IFEval, MBPP & 22,991 \\
\bottomrule
\end{tabular}
}
\end{table}

\subsection{Benchmarks for Offline Reinforcement Learning Evaluation}
\label{sec:dpo}
To comprehensively assess the policy derived from DPO, we conduct evaluations across two distinct domains: mathematical reasoning and open-ended instruction following.

\paragraph{Mathematical Reasoning.} We employ a suite of four challenging datasets to evaluate the model's deductive logic and problem-solving capabilities: GSM8k \citep{cobbe2021training}, MATH \citep{hendrycks2021measuring}, MAWPS \citep{koncel2016mawps}, and TabMWP \citep{lu2023dynamic}. These benchmarks cover a wide spectrum of difficulty, ranging from grade-school arithmetic to competition-level mathematics and tabular processing.

\paragraph{Instruction Following.} For general alignment and conversational versatility, we utilize two widely adopted benchmarks: AlpacaEval-2 \citep{dubois2024lengthcontrolled} and Arena-Hard v0.1 \citep{arenahard2024}. Evaluation is performed using an auto-evaluator in a head-to-head setting, where the model's responses are compared against a baseline reference to determine win rates. We strictly adhere to the officially recommended configurations for reproducibility.

\subsection{Benchmarks for Test-time Scaling Evaluation}
\label{sec:tts}
Following the \textbf{JETTS} setup \citep{zhou2025evaluating}, we perform Best-of-10 reranking evaluations where the model selects the optimal solution from a mixed pool of candidate responses. We report results on the four most challenging subsets of the benchmark: \textbf{MATH} \citep{hendrycks2021measuring} for mathematical reasoning, \textbf{CHAMP} \citep{mao2024champ} for competition-level math, along with \textbf{MBPP+} \citep{liu2023is} and \textbf{BigCodeBench} \citep{zhuo2025bigcodebench} for code generation. This selection tests the model's ability to identify correct reasoning paths in complex scenarios.

\section{Prompts Template}

To align with established community standards, our Vanilla-CoT generation employs the representative prompts originally introduced in MT-Bench~\citep{zheng2023judging} and RewardBench~\citep{lambert2024rewardbench}.
\begin{figure*}[t]
\begin{promptbox}{Prompt for Vanilla-CoT Generation}
Please act as an impartial judge and evaluate the quality of the responses provided by two AI assistants to the user question displayed below. You should choose the assistant that follows the user's instructions and answers the user's question better. Your evaluation should consider as many factors as possible. Begin your evaluation by comparing the two responses and provide a through reasoning. Avoid any position biases and ensure that the order in which the responses were presented does not influence your decision. Do not allow the length of the responses to influence your evaluation. Do not favor certain names of the assistants. Be as objective as possible. After providing your reasoning, output your final verdict by strictly following this format: \"[[A]]\" if assistant A is better, \"[[B]]\" if assistant B is better. 

[Instruction]

{instruction}

[The Start of Assistant A's Answer]

\{response\_a\}

[The End of Assistant A's Answer]

[The Start of Assistant B's Answer]

\{response\_b\}

[The End of Assistant B's Answer]
\end{promptbox}
\end{figure*}
Upon generating the raw Vanilla-CoT using the standard prompts, we employ a specialized extraction prompt to parse the unstructured text into the modular ``Principle–Judgment–Verdict‘' schema.
\begin{figure*}[t]
\begin{promptbox}{Prompt for Schema Extraction}
PRIMARY TASK:

Your mission is to analyze a given reasoning Chain-of-Thought from a generative reward model. From this CoT, you will extract, define, and refine the specific, detailed principles (or rubrics, criteria) it uses to judge the quality of AI-generated responses. For each principle, you must provide a corresponding analysis that traces it directly back to the original text.

INSTRUCTIONS:

You will be given a CoT text below. Please follow these four steps precisely:

1. Deconstruct the CoT: First, perform a close reading of the entire CoT. Identify all explicit evaluation criteria mentioned as well as any implicit judgments or preferences revealed in the model's comparative language.

2. Extract the Core Idea of Each Criterion: For each criterion, do not simply use the high-level category name. Your goal is to uncover the specific description of that criterion as used by the model. Ask yourself: What specific actions, qualities, or content does the model praise or criticize? What makes one response ``more accurate" or ``clearer" according to this specific CoT?

3. Formulate and Refine the Principle: Convert each core idea you extracted into a formal, normative, and reusable principle.

3.1  Name It: Give the principle a clear and descriptive name that captures its essence (e.g., ``Principle of Factual Precision," ``Principle of Structural Clarity").

3.2 Define It: Write the principle as a concise, actionable, and universal rule. It should be an instructive statement about what constitutes a high-quality response.

3.3 Be Specific: Avoid vague terms. Instead of ``The response should be relevant," specify how it should be relevant based on the CoT's logic, such as ``A relevant response must directly and unambiguously address the user's primary question."

3.4 Be Normative: Phrase it as a standard to be met (e.g., ``A high-quality response must...").

4. Provide Corresponding Judgment: For each principle you formulate, you must write a brief ``CoT Judgment Extraction." To do this, quote or closely paraphrase specific phrases from the CoT that support your formulation.

5. Conclude the sub-verdict in this Judgment: For the principle and corresponding judgment, you should conclude this verdict in this segment.

OUTPUT FORMAT:

You must follow this format strictly for your entire response.

. . .

\#\#\# 1. Principle of [Descriptive Name]: [Your refined, normative principle statement.]
\textbf{Judgment:} [In this principle, what judgment on Response A and Response B quotes or paraphrases from the source CoT.]
\textbf{Sub-Verdict:} <<A/B>>, In this principle, the judgment judge which assistant Better]

\#\#\# 2. Principle of [Descriptive Name]: [Your refined, normative principle statement.]
\textbf{Judgment:} [In this principle, what judgment on Response A and Response B quotes or paraphrases from the source CoT.]***
\textbf{Sub-Verdict:} <<A/B>>, In this principle, the judgment judge which assistant Better]

(Continue this structure for all principles identified in the CoT)

. . .

Extract and Analyse the following CoT Text

\{Vanilla-CoT\}
\end{promptbox}
\end{figure*}
Leveraging the parsed schemas, we introduce specialized prompts to synthesize the two target morphologies. For Breadth-CoT, the synthesis process entails merging modular components derived from at least two Vanilla-CoT responses, followed by a deduplication step to ensure diverse coverage.
\begin{figure*}[t]
\begin{breadthprompt}{Prompt for Breadth-CoT Generation}
PRIMARY TASK:

You are provided with a series of lists, each containing Principles, Judgments, and Sub-Verdicts derived from an independent analysis of a Chain-of-Thought (CoT). Your mission is to merge these lists into a single, master list of unique evaluation principles.

INSTRUCTIONS FOR MERGING AND SYNTHESIS:

1.  \textbf{Deduplication and Semantic Grouping:}

    * Compare all Principles with the corresponding Judgments across all provided lists.
    * Identify and group principles that are \textbf{semantically similar}, even if they use different wording (e.g., ``Principle of Precision" and ``Principle of Correcteness" are likely the same concept).

2.  \textbf{Principle Refinement:}

    * For each semantic group, synthesize the most concise, actionable, and specifically-detailed statement for the \textbf{Principle Description}.
    * Select the most descriptive and formal \textbf{Name} for the refined principle.

3.  \textbf{Judgment Synthesis:}

    * For the refined principle, create a new, synthesized \textbf{Judgment} block. This block should consist of a curated selection of the most illustrative quotes and paraphrases from the original Judgments across all source lists that led to the consolidated principle. This new Judgment serves as the combined evidence for the principle.

4.  \textbf{Merge Count:}

    * \textbf{COUNT THE SOURCES:} For each synthesized principle, you must count the total number of original, distinct principles/judgments from the source lists that were merged to create it. This number is the \textbf{Merge Count}.

5.  \textbf{Sub-Verdict Aggregation:}

    * The final \textbf{Sub-Verdict} for the synthesized principle must reflect the aggregated trend. Since the judgments are now synthesized, simply use the majority verdict (e.g., if a principle appeared 4 times with [[B]] and 1 time with [[A]], conclude [[B]]). If the verdicts are balanced (e.g., 2 [[A]] and 2 [[B]]), state \textbf{[[MIXED]]}.

6.  \textbf{Strict Output Adherence:}

    * Maintain the exact four-part format for every final entry. The output must be one continuous list of unique, synthesized principles.

SOURCE LISTS TO MERGE:

[Insert List 1 Here]

[Insert List 2 Here]

[Insert List 3 Here]

(Continue for all lists)

OUTPUT FORMAT:

You must follow this exact format for your final, merged response.

***

\#\#\# 1. Principle of [Refined, Descriptive Name]: [The synthesized, normative principle statement.]
\textbf{Judgment:} [A synthesis of the most relevant quotes/paraphrases from the source Judgments that supports this consolidated principle.]***
\textbf{Merge Count:} [The total number of original source principles/judgments that were merged to form this entry.]
\textbf{Sub-Verdict:} <<A/B/MIXED>>, The aggregate verdict for this principle across all CoTs.]

\#\#\# 2. Principle of [Refined, Descriptive Name]: [The synthesized, normative principle statement.]
\textbf{Judgment:} [A synthesis of the most relevant quotes/paraphrases from the source Judgments that supports this consolidated principle.]***
\textbf{Merge Count:} [The total number of original source principles/judgments that were merged to form this entry.]
\textbf{Sub-Verdict:} <<A/B/MIXED>>, The aggregate verdict for this principle across all CoTs.]

(Continue this structure for all unique, synthesized principles)

***

\end{breadthprompt}
\end{figure*}
In contrast, the synthesis of Depth-CoT relies on a reasoning-guided evaluation mechanism. We initially prompt the model to reason the instruction deeply. We then use this generated reasoning to ground the re-assessment of selected principles extracted from the parsed schemas, discarding their previous rationales to ensure the new judgments are purely driven by rigorous reasoning.
\begin{figure*}[t]
\begin{depthprompt}{Prompt for Depth-CoT Verification}
PRIMARY TASK:
Your role is to critically assess the quality of two competing responses (Assistant A and Assistant B) against the user's question, leveraging the expert reasoning as the ultimate ground truth.

\textbf{MANDATORY NON-BIAS RULES:} Avoid all position biases (do not favor the first response presented). Do not allow the length or formatting of the responses to influence your evaluation. Do not favor certain names of the assistants. Be as objective, clinical, and data-driven as possible.

\textbf{PRINCIPLE-BASED EVALUATION}

Given some potential principles, \{principles\}, you should choose the most critical principle (Preferably one principle, with a maximum of three) from them and then evaluate the two Chatbot responses (A and B) based on the choosed principle. This evaluation must directly reference the deep reasoning to instruction and **\textbf{must strictly adhere to the following output format for each principle:}**

EXPERT REASONING: \{reasoning\}

\#\#\# Principle of [Critical Principle Name]: 

\textbf{Judgment:} [Give your specific and detailed evaluation in this principle, and if you are referring the this reasoning, you **MUST** quote using `<Answer>`]

\textbf{Sub-Verdict:} <<A/B/MIXED>>, <<A>> if assistant A is better in this principle, <<B>> if assistant B is better in this principle, <<MIXED>> if assistant A and B is Tie.

After providing your complete principle-based evaluation, output your final verdict by strictly following this format: \"[[A]]\" if assistant A is better, \"[[B]]\" if assistant B is better.
\end{depthprompt}
\end{figure*}
\section{Reward Model Performance Across Preference and Correctness}
\label{sec:performance_preference_correctness}
To provide a more granular view of our model's efficacy, we report detailed performance across specific tasks based on the meta-data provided by each benchmark. We categorize these tasks into two distinct tables: Table~\ref{tab:preference_performance} for subjective preference tasks and Table~\ref{tab:correctness_performance} for objective correctness tasks. This fine-grained reporting serves as a detailed decomposition of the mechanism-level performance discussed in the main text, offering deeper empirical evidence for the mechanism-task synergy between \textsc{B-CoT} and \textsc{D-CoT}.

\begin{table*}[ht]
\caption{Performance of RMs on preference-related sub-tasks. ``Avg.'' is the average score among sub-tasks. Best per column is \textbf{bolded}; second-best is \underline{underlined}.}
\label{tab:preference_performance}
\begin{center}
\resizebox{.96\textwidth}{!}{%
\begin{tabular}{l c cc c c cc c} % 1 + 8 = 9 columns (visually grouped)
\toprule
\textbf{Models} &
\makecell{\textbf{Reward}\\\textbf{Bench}} &
\multicolumn{2}{c}{\makecell{\textbf{Reward}\\\textbf{Bench-v2}}} &
\makecell{\textbf{RM}\\\textbf{Bench}} &
\textbf{RMB} &
\multicolumn{2}{c}{\textbf{PPE}} &
\textbf{Avg.} \\
\cmidrule(lr){2-2} \cmidrule(lr){3-4} \cmidrule(lr){5-5} \cmidrule(lr){6-6} \cmidrule(lr){7-8}
& \textbf{\textsc{Chat}} & \textbf{\textsc{Focus}} 
& \textbf{\textsc{IF}} & \textbf{\textsc{Chat}}
& \textbf{\textsc{Helpfulness}}& \textbf{\textsc{Human}}
& \textbf{\textsc{IF}}& \\
\midrule
\multicolumn{9}{l}{\underline{\textit{Open-sourced Reward Models}}} \\
%\rowindent Skywork-Reward-Llama-3.1-8B & 91.3& 93.8& 40.6& 68.8 & 75.6& 62.0& 61.3&70.5\\
\rowindent JudgeLRM-7B                  &75.8 &46.8 &31.3 &68.4 &\underline{79.3} &60.2 &54.3 &59.4 \\
\rowindent RM-R1-7B (Distill)           & 75.3& 58.7& 24.4& 58.7& 63.2& 57.1& 53.0& 55.8\\
\rowindent RM-R1-7B (Instruct)          & 80.5& 85.3& 28.8& 64.7& 65.1& 57.1& 53.0& 62.1\\
\rowindent FARE-8B                      &85.0 & 78.4&\underline{36.3} &66.9 & \textbf{82.9}&63.4 &55.7 &67.0 \\
\rowindent RubricRM-8B                  &82.4 & 78.2& 33.8& 62.2& 77.5&63.8 &\textbf{66.0} & 66.3\\
\rowindent DeepSeek-GRM-16B             &80.6 & 79.2& 40.0& 64.0& 76.8&61.7 &57.9 & 65.7\\
\midrule
\multicolumn{9}{l}{\underline{\textit{Our Proposed Reward Models}}} \\
    \rowcolor{blue!12}
    \multicolumn{9}{l}{\textbf{SFT-trained}} \\
    \rowindent Base-GRM   & 81.6& 76.6& 34.4& 63.3& 80.5& 64.3& 55.9& 65.2\\
    \rowcolor{gray!12}
    \rowindent Mix-GRM (Breadth)  & 83.7& 84.5& 33.8& 65.9& 77.9& 63.5& 55.6& 66.4\\
    \rowindent Mix-GRM (Depth)    & 80.3& 72.2& 28.1& 70.6& 70.1& 63.1& 54.1& 62.6\\
    \rowcolor{gray!12}
    \rowindent Mix-GRM   & 84.9& 79.6& 31.9& 71.2& 78.7& 62.0& 56.3& 66.4\\
    \addlinespace[1pt]
    \rowcolor{green!12}
    \multicolumn{9}{l}{\textbf{RLVR-trained}} \\
    \rowindent Base-GRM   & 83.0& 86.7& 29.4& 68.5& 73.8& 65.8& 57.1& 66.3\\
    \rowcolor{gray!12}
    \rowindent Mix-GRM (Breadth)  &\textbf{86.2}	&88.9 &28.8 &70.1 &79.2&\textbf{66.0}&55.3 &\underline{67.8}\\
    \rowindent Mix-GRM (Depth)    &\underline{85.3}&\underline{89.3} &26.3&\textbf{75.6}&75.4&\textbf{66.0}&	56.4&\underline{67.8} \\
    \rowcolor{gray!12}
    \rowindent Mix-GRM   & \textbf{86.2}& \textbf{91.3}&\textbf{37.5}&\underline{72.7}&78.1&\underline{65.9}&\underline{57.4}&	\textbf{69.9}\\
\bottomrule
\end{tabular}
}
\end{center}
\end{table*}

\begin{table*}[ht]
\caption{Performance of RMs on correctness-related sub-tasks. ``Avg.'' is the average within this block. Best per column is \textbf{bolded}; second-best is \underline{underlined}.}
\label{tab:correctness_performance}
\begin{center}
\resizebox{\textwidth}{!}{%
\begin{tabular}{l c cc c c cc c cc cc c} % 1 + 8 = 9 columns (visually grouped)
\toprule
\textbf{Models} &
\multicolumn{2}{c}{\textbf{RewardBench}} &
\multicolumn{2}{c}{\textbf{RewardBench-v2}} &
\multicolumn{2}{c}{\textbf{RM-Bench}} &
\textbf{RMB} &
\multicolumn{4}{c}{\textbf{PPE}} &
\textbf{Avg.} \\
\cmidrule(lr){2-3} \cmidrule(lr){4-5} \cmidrule(lr){6-7} \cmidrule(lr){8-8} \cmidrule(lr){9-12}
& \textbf{\textsc{Code}} & \textbf{\textsc{Math}} 
& \textbf{\textsc{Factuality}} & \textbf{\textsc{MATH}}
& \textbf{\textsc{CODE}}& \textbf{\textsc{MATH}}
& \textbf{\textsc{CODE}}& \textbf{\textsc{MMLU-Pro}} & \textbf{\textsc{MATH}}& \textbf{\textsc{GPQA}}& \textbf{\textsc{MBPP}} &\\
\midrule
\multicolumn{13}{l}{\underline{\textit{Open-sourced Reward Models}}} \\
%\rowindent Skywork-Reward-Llama-3.1-8B & 95.3& 97.5& 69.0& 59.0& 53.2& 61.9&84.6 & 60.5&71.5&55.8&58.6&69.7\\
\rowindent JudgeLRM-7B                  & 81.6& 77.2& 53.8& 76.5& 51.0& \textbf{86.7}& 82.1& 57.2 &65.5&51.3&52.3&66.8\\
\rowindent RM-R1-7B (Distill)           & 91.9&\textbf{93.7}&28.3&73.2&53.3&\underline{85.8}&74.8&66.7&\textbf{89.4}&\textbf{56.3}&\underline{64.4}&70.7\\
\rowindent RM-R1-7B (Instruct)          &81.7&84.1&42.6&67.8&56.7&72.7&74.7&\textbf{67.0}&\underline{89.1}&\underline{55.9}&\textbf{64.8}&68.8\\
\rowindent FARE-8B                      & 88.1& 82.3& \textbf{65.8}& 68.9& 57.0& 69.1& 88.1& 63.2&79.3&55.2&55.4&70.2\\
\rowindent RubricRM-8B                  & 93.6& 81.7& 50.8& 77.6& 55.4& 59.8& 86.5& 60.9&75.5&52.8&52.4&67.9\\
\rowindent DeepSeek-GRM-16B             &84.0 & 69.1& 49.4& 62.3& 51.5&61.7 &86.8 & 55.2 &64.3&54.1&53.7&62.9\\
\midrule
\multicolumn{13}{l}{\underline{\textit{Our Proposed Reward Models}}} \\
    \rowcolor{blue!12}
    \multicolumn{13}{l}{\textbf{SFT-trained}} \\
    \rowindent Base-GRM   & 91.0& 77.2& 46.0& 81.4& 57.1& 78.4& 86.4& 59.9 &71.8 &52.2 &52.5&68.5\\
    \rowcolor{gray!12}
    \rowindent Mix-GRM (Breadth)  & 90.1& 72.0& 51.1& 69.4& 54.0& 74.1& 86.8&60.0&70.3&51.9&52.4&66.6 \\
    \rowindent Mix-GRM (Depth)    & 89.8& 86.1& 45.1& 75.4& 56.2& 77.1& 81.1& \underline{66.8}&83.6&54.7&53.7&70.0\\
    \rowcolor{gray!12}
    \rowindent Mix-GRM   &88.9&87.9&55.7&76.0&55.8&	79.6&81.9&63.9&	82.1 &54.8 &54.0& 71.0 \\
    \addlinespace[1pt]
    \rowcolor{green!12}
    \multicolumn{13}{l}{\textbf{RLVR-trained}} \\
    \rowindent Base-GRM   & 93.2 &86.4&\underline{62.0}&77.0 &61.7&78.1&\textbf{89.5}&64.6&84.1&54.3&	50.5&72.9\\
    \rowcolor{gray!12}
    \rowindent Mix-GRM (Breadth)  & \textbf{96.3} & 69.3 & 55.7 & 71.0 &58.0	&70.6& 86.5& 61.7& 75.2 &53.0& 52.8&	68.2 \\
    \rowindent Mix-GRM (Depth)    & 94.6 &\underline{89.0}& 61.8&\underline{78.7}& \underline{64.4}&81.4 & 87.4 & \textbf{67.0} & 86.5 &	55.6 &55.5 &\underline{74.7} \\
    \rowcolor{gray!12}
    \rowindent Mix-GRM   & \underline{95.4}	&\underline{89.0}&	\textbf{65.8}& \textbf{79.2}&	\textbf{66.6}& 82.5 & \underline{88.9} &	65.0 & 86.7 & 54.8 & 55.2 &	\textbf{75.4} \\
\bottomrule
\end{tabular}
}
\end{center}
\end{table*}

\section{Sensitivity and Robustness}
\label{sec:sensitivity}

The data synthesis process consists of two primary phases: the sampling of raw rationales and the subsequent synthesis of B-CoT/D-CoT. To guarantee the reliability of this pipeline, we conduct empirical analyses focusing on pipeline stability and robustness to noise.

\paragraph{Quantitative Evaluation of Pipeline Stability.}
To ensure our extraction and refinement modules do not introduce systematic errors or degrade data quality, we tracked the solution accuracy of the synthesized data at each pipeline stage on the training set. As shown in Table~\ref{tab:pipeline_stability}, the accuracy remains highly stable across the transformations. The merge step (B-CoT) resolves contradictions and improves accuracy, while the D-CoT generation maintains high fidelity to the correct reasoning paths. This confirms that the intermediate processing steps reliably preserve data quality without catastrophic degradation.

\begin{table}[h]
    \centering
    \small
    \resizebox{.48\textwidth}{!}{
    \begin{tabular}{lccc}
        \toprule
        \textbf{Synthesis Stage} & Raw Rationale & Merge B-CoT & Generate D-CoT \\
        \midrule
        \textbf{Accuracy (\%)} & 87.1 & 90.2 & 88.5 \\
        \bottomrule
    \end{tabular}}
    \caption{Solution accuracy tracked across different stages of the synthesis pipeline, demonstrating the stability of the intermediate transformations.}
    \label{tab:pipeline_stability}
\end{table}

\paragraph{Robustness to Noise.}
A common concern with LLM-generated rationales is the necessity of strict filtering to ensure perfect correctness. We conducted an ablation study during the SFT stage to measure the impact of noise. We compared a model trained on $9$K strictly verified CoT data (filtered for correct final verdicts) against one trained on $9$K noisy CoT data (without strict correctness filtering). As shown in Table~\ref{tab:noise_robustness}, the SFT performance difference is negligible. This indicates that our method is highly robust to noise in the synthesized rationales and does not require perfect accuracy from the extraction pipeline to be effective. Consequently, we adopt the unfiltered data setting for our main experiments to significantly reduce the computational cost of data curation without sacrificing performance.

\begin{table}[h]
    \centering
    \small
    \resizebox{.48\textwidth}{!}{
    \begin{tabular}{lcc}
        \toprule
        \textbf{Training Data (9K)} & w/o Filtering (Noisy) & w/ Filtering (Verified) \\
        \midrule
        \textbf{SFT Performance (\%)} & 69.8 & 70.1 \\
        \bottomrule
    \end{tabular}}
    \caption{SFT performance comparison between noisy and strictly filtered synthesized data.}
    \label{tab:noise_robustness}
\end{table}

Furthermore, we utilize the open-weights \textsc{DeepSeek-V3} model for both schema extraction and raw rationale generation. The effectiveness of our synthesis pipeline with an open-source model further underscores its robustness and generalizability, proving it is not overly sensitive to the choice of the underlying LLM.

\end{document}